\documentclass{article}
\usepackage[final]{neurips_2026}

\makeatletter
\def\@trackname{NeurIPS 2026 Workshop}
\makeatother

\usepackage[utf8]{inputenc} 
\usepackage[T1]{fontenc}    
\usepackage{hyperref}       
\usepackage{url}            
\usepackage{booktabs}       
\usepackage{amsfonts}       
\usepackage{nicefrac}       
\usepackage{microtype}      
\usepackage[table]{xcolor}         
\usepackage{array}
\usepackage{tabularx}
\usepackage{makecell}
\usepackage{booktabs}
\usepackage{graphicx}
\usepackage{subcaption}
\usepackage{multirow}
\usepackage{tcolorbox}
\usepackage{float}
\usepackage{placeins}
\tcbuselibrary{listings,skins,breakable}
\usepackage{xcolor}
\newcolumntype{L}{>{\raggedright\arraybackslash}X}
\newcolumntype{C}{>{\centering\arraybackslash}X}
\makeatletter
\renewcommand{\footnoterule}{%
  \kern -3pt
  \hrule width \textwidth height 0.4pt
  \kern 2.6pt
}
\makeatother

\usepackage{etoolbox}

\renewenvironment{abstract}
{
  \vskip 0.075in
  \centerline{\large\bf Abstract}
  \vspace{0.5em}
  \noindent
  \begin{minipage}{\textwidth}
}
{
  \end{minipage}
  \vskip 0.075in
}

\usepackage[utf8]{inputenc}   
\usepackage[T1]{fontenc}      

\usepackage{amsmath}          
\usepackage{amsfonts}         
\usepackage{amssymb}          
\usepackage{nicefrac}         

\usepackage{booktabs}         
\usepackage{microtype}        
\usepackage{xcolor}           
\usepackage{enumitem}         
\usepackage{listings}

\usepackage{url}              
\usepackage{hyperref}         

\usepackage{tikz}             
\usetikzlibrary{
    positioning,              
    arrows.meta,              
    shapes.geometric          
}

\usepackage{pgfplots}         
\pgfplotsset{compat=1.18}     

\lstdefinestyle{jsonstyle}{
    basicstyle=\ttfamily\scriptsize,
    breaklines=true,
    frame=single,
    columns=fullflexible,
    keepspaces=true,
    showstringspaces=false,
    tabsize=2
}

\title{PDEFlow: Autonomous Agentic PDE Pipelines for Neural Operator Learning and Solver-Free Inference}

\author{%
\makebox[0pt][c]{%
\begin{minipage}{\textwidth}
\centering
Akshat Jani\textsuperscript{*}
\hspace{0.65em}
Prathamesh Gadekar\textsuperscript{*}
\hspace{0.65em}
Sakhinana Sagar Srinivas
\hspace{0.65em}
Venkataramana Runkana
\\[0.45em]
{\normalfont\mdseries Tata Research Development and Design Center, Pune, India 411057}
\end{minipage}%
}%
}

\begin{document}

\maketitle

\begingroup
\renewcommand{\thefootnote}{}
\footnotetext{%
\noindent
\textsuperscript{*}Equal Contribution. Correspondence to: Sakhinana Sagar Srinivas <sagar.sakhinana@tcs.com>.
}
\endgroup


\begin{abstract}
We present PDEFlow, an autonomous agentic framework that turns user-level ODE and PDE descriptions into solver-backed neural-operator pipelines. The workflow links problem specification, data generation, operator training, and checkpoint-based inference. A stateful input graph converts multi-turn natural-language input and user edits into validated problem specifications. The data-generation module then samples parameters, solves the configured governing-equation with FEniCSx finite-element backend, and stores the solutions as operator-ready tensors. The training and inference stages use a registry-based interface, allowing different neural operators to be trained and deployed without changing the surrounding pipeline. In the current implementation, we instantiate this interface with a multi-branch Bayesian DeepONet. Experiments on benchmark ODE and PDE tasks show that PDEFlow can construct valid specifications, generate solver-backed datasets, train neural operators across steady and transient problem classes, and provide solver-free predictions from saved checkpoints. The framework is designed for repeatable scientific and engineering workflows where many related physics configurations must be specified, simulated, learned, and queried with minimal manual intervention.
\end{abstract}


\section{Introduction}
\label{sec:introduction}

Many scientific design tasks involve repeated evaluation of the same governing physics under changing conditions. An airfoil is tested across angles of attack and geometric variants; a heat exchanger is studied under different material parameters; a reactor model is run across sampled coefficients for uncertainty quantification. The pattern is simple. The same equation is solved many times.

This repetition defines downstream workflows such as design optimisation, uncertainty quantification and design-space exploration. Each query requires a valid specification, a solver run, and post-processing. The cost is high. Not only in computation, but also in the process of turning a researcher’s description into an executable form. That step is often manual. It is also fragile.

A user rarely writes a complete specification in one attempt. The equation may come first. Boundary conditions are corrected later. Coefficient distributions are added after initial runs. Small edits accumulate. A wrong boundary edge or an outdated parameter can silently affect all subsequent results. These inconsistencies are easy to miss because most workflows treat every update as a fresh configuration rather than a controlled modification of state.

Neural operators address the cost of repeated solves. Once trained, a model can approximate solutions over a range of inputs without calling the numerical solver. This is useful when exploring many configurations. But operator learning usually begins after two difficult steps have already been completed: the problem specification has been made valid, and the solver-backed dataset has been generated. The earlier workflow remains unresolved. Can an agentic system reliably construct, revise, and audit differential-equation specifications across multiple user interactions? Can the same workflow automatically generate solver-backed training data, train a neural operator, and reuse the saved checkpoint for solver-free inference? More broadly, can natural-language problem entry be connected to an end-to-end pipeline for specification, simulation, learning, and fast prediction?

We introduce a system that connects specification, data generation, model training and inference in a single workflow. The system converts multi-turn natural-language input into a validated JSON specification, generates solver-backed datasets, trains neural operators, and performs inference from saved checkpoints without re-solving the equations. The design is stateful. Each user edit is treated as a local modification, expressed through JSON patches that are validated and, if needed, corrected before being applied. This prevents silent drift when equations, coefficients, or boundary conditions change across turns.

The resulting workflow supports end-to-end automation of physics-based pipelines. From a user description to predictions. With intermediate artifacts preserved. The aim is not to replace numerical solvers, but to make repeated simulation workflows easier to specify, reuse and trust when exploring large design spaces.


\section{Related Works}
\label{sec:related_work}

Scientific simulation workflows usually rely on numerical solvers first. Finite-element tools such as FEniCSx \citep{baratta2023dolfinx,scroggs2022basixdof,scroggs2022basix,alnaes2014ufl,fenicsproject2025fenicsx010} and UFL support weak-form specification and assembly \citep{logg2012fenics,alnaes2014ufl}, while PETSc provides sparse linear algebra and time-stepping support for solver pipelines \citep{balay2026petsc}. Once a valid configuration exists, these tools are effective. The harder step often comes before the solver call: users must translate equations, coefficients, domains, initial conditions and boundary conditions into an executable form, sometimes across several edits.

Operator learning reduces the cost of repeated solves after this setup is complete. PINNs train networks using equation residuals and data terms \citep{raissi2019physics}; neural operators instead learn solution maps across functions or parameters. DeepONet uses a branch--trunk architecture \citep{lu2021deeponet}, Fourier neural operators use spectral layers for grid-based problems \citep{li2021fourier}, and \citet{kovachki2023neural} review neural-operator theory and applications. MIONet extends this setting to multiple input functions \citep{jin2022mionet}. Bayesian neural networks and Bayesian DeepONet variants add predictive spread for design exploration \citep{blundell2015weight,garg2022variational}. These methods help once training data exist, but they do not address how the solver-ready specification and dataset are created.

Agentic systems target this missing workflow layer. ReAct links reasoning with tool actions \citep{yao2023react}, AutoGen supports multi-agent interaction \citep{wu2023autogen}, and LangGraph provides stateful graph execution \citep{langgraph2026}. Scientific agents such as ChemCrow, Coscientist and AutoChemSchematic AI connect language models with domain tools and simulators \citep{bran2024chemcrow,boiko2023coscientist,srinivas2025autochemschematic}. PDEFlow follows this direction for solver-backed operator learning: it keeps a differential-equation specification valid across user edits, generates finite-element data from the accepted specification and trains a registered neural operator for solver-free inference.


\section{Methodology}
\label{sec:methodology}

\begin{figure*}[htbp]
    \centering
    \includegraphics[width=\textwidth]{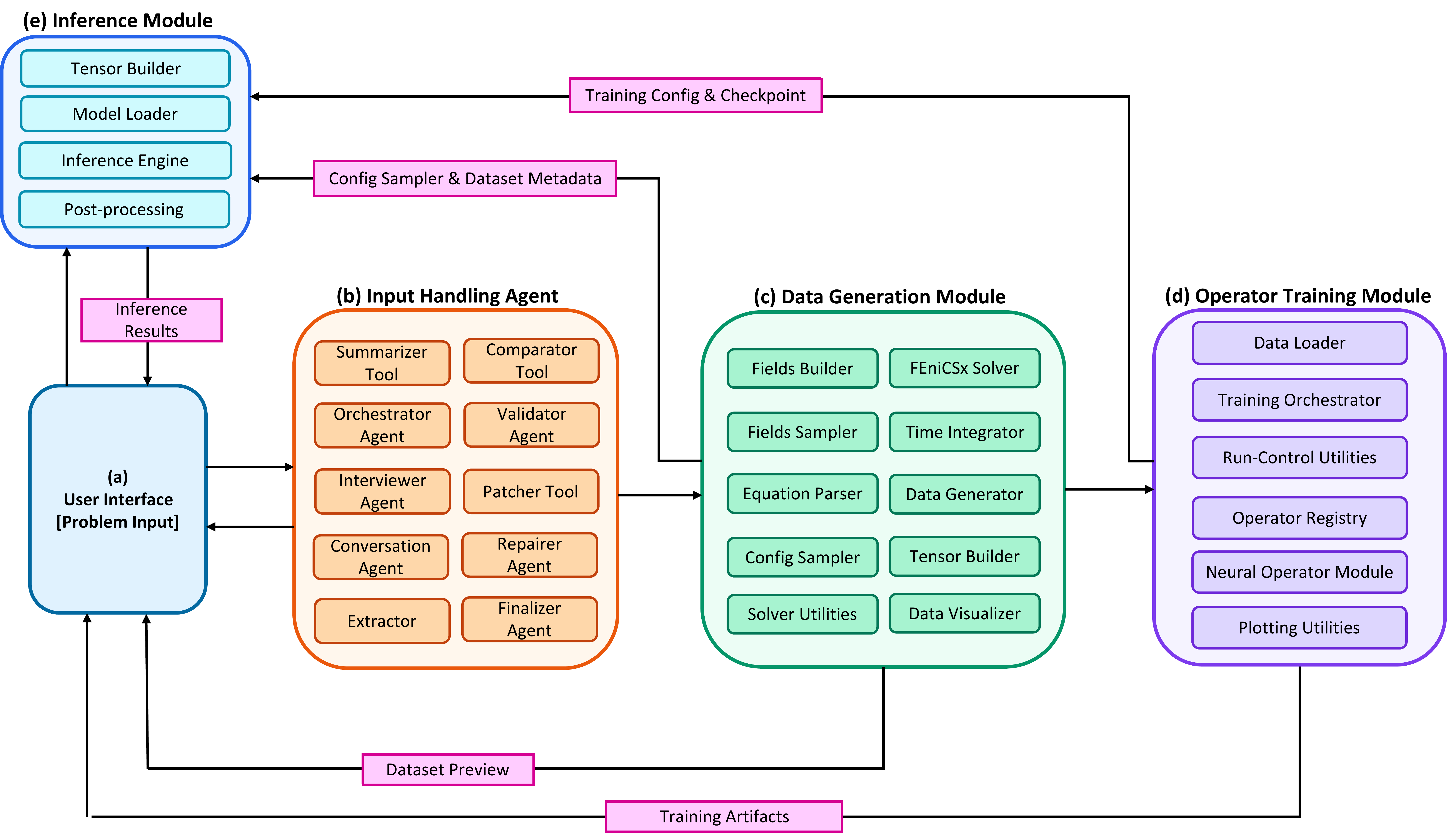}
    \captionsetup{justification=justified,singlelinecheck=false}
    \caption{Complete PDEFlow system architecture. A user-provided problem description enters through the user interface (a) and is converted by the input-handling agent (b) into a validated executable specification. The data-generation module (c) parses the specification, samples coefficients and fields, solves the configured problem with FEniCSx, and builds operator-ready tensors. The operator-training module (d) trains a registered neural operator and saves training artifacts, while the inference module (e) reuses the saved configuration, checkpoint, sampled configuration, and dataset metadata to generate solver-free prediction results.}
    \label{fig:complete_system_architecture}
\end{figure*}


\subsection{Problem Formulation}
\label{sec:problem_formulation}

We consider linear one- and two-dimensional physics problems defined on
$\Omega \subset \mathbb{R}^d$, $d\in\{1,2\}$, with boundary
$\partial\Omega=\Gamma_L\cup\Gamma_R\cup\Gamma_B\cup\Gamma_T$ in the 2D case.
The solver receives a PDE configuration consisting of coefficient fields,
forcing terms, initial conditions, and edge-wise boundary conditions, and returns
a steady solution $u(x,y)$ or a transient solution $u(t,x,y)$.

The general transient PDE supported by the solver is
\begin{equation}
    \frac{\partial u}{\partial t}
    =
    \nabla\cdot\left(K(\mathbf{x})\nabla u\right)
    + \boldsymbol{\alpha}(\mathbf{x})\cdot\nabla u
    + \gamma(\mathbf{x})u
    + f(\mathbf{x},t),
    \qquad \mathbf{x}\in\Omega,\; t\in[0,T],
    \label{eq:general_pde}
\end{equation}
with initial condition
\begin{equation}
    u(\mathbf{x},0)=u_0(\mathbf{x}).
\end{equation}
Here,
\begin{equation}
    K(\mathbf{x})=
    \begin{bmatrix}
        \beta_x(\mathbf{x}) & \eta_{xy}(\mathbf{x})\\
        \eta_{yx}(\mathbf{x}) & \beta_y(\mathbf{x})
    \end{bmatrix},
    \qquad
    \boldsymbol{\alpha}(\mathbf{x})=
    \begin{bmatrix}
        \alpha_x(\mathbf{x})\\
        \alpha_y(\mathbf{x})
    \end{bmatrix},
\end{equation}
where $K$ is the anisotropic diffusion tensor, $\boldsymbol{\alpha}$ is the
advection field, $\gamma$ is the reaction coefficient, and $f$ is the source
term. For steady-state simulations, the time derivative in
Eq.~\eqref{eq:general_pde} is omitted.

The solver supports edge-wise boundary conditions. For an edge
$\Gamma_e$, $e\in\{L,R,B,T\}$, the boundary condition can be specified as
\begin{align}
    \text{Dirichlet:} \qquad
    & u(\mathbf{x},t)=u_e(\mathbf{x},t),
    && \mathbf{x}\in\Gamma_e, \\
    \text{Neumann:} \qquad
    & -\mathbf{n}\cdot K\nabla u = g_e(\mathbf{x},t),
    && \mathbf{x}\in\Gamma_e, \\
    \text{Robin:} \qquad
    & a_e(\mathbf{x},t)u
      + b_e(\mathbf{x},t)\left(-\mathbf{n}\cdot K\nabla u\right)
      = c_e(\mathbf{x},t),
    && \mathbf{x}\in\Gamma_e,
\end{align}
where $\mathbf{n}$ is the outward unit normal. Different boundary types may be
assigned independently to different edges.

The numerical solver defines a solution operator
\begin{equation}
    \mathcal{G}^{\dagger}: a \mapsto u,
    \qquad
    a=\left(u_0,f,K,\boldsymbol{\alpha},\gamma,\mathcal{B},\Omega,T\right),
\end{equation}
where $\mathcal{B}$ denotes the boundary-condition specification. The goal of
operator learning is to approximate $\mathcal{G}^{\dagger}$ with a neural
operator
\begin{equation}
    \mathcal{G}_{\phi}:
    \left(X_f,X_s,z\right)\mapsto \hat{u}(z),
\end{equation}
where $X_f$ contains functional inputs such as initial conditions, coefficients,
forcing fields, or boundary encodings; $X_s$ contains scalar simulation
parameters; and $z$ is a spatial or spatio-temporal query point. For steady
problems $z=(x,y)$, while for transient problems $z=(t,x,y)$.

Given solver-generated data
\begin{equation}
    \mathcal{D}
    =
    \left\{
    \left(X_f^{(i)},X_s^{(i)},z^{(i)},y^{(i)}\right)
    \right\}_{i=1}^{N},
\end{equation}
the operator is trained by minimizing
\begin{equation}
    \phi^{*}
    =
    \arg\min_{\phi}
    \frac{1}{N}
    \sum_{i=1}^{N}
    \left\|
    \mathcal{G}_{\phi}
    \left(X_f^{(i)},X_s^{(i)},z^{(i)}\right)
    -
    y^{(i)}
    \right\|_2^2 .
\end{equation}
At inference time, the trained checkpoint $\phi^{*}$ is restored and the
operator predicts
\begin{equation}
    \hat{u}_{n}
    =
    \mathcal{G}_{\phi^{*}}
    \left(
    X_f^{\mathrm{inf}},
    X_s^{\mathrm{inf}},
    Z^{\mathrm{inf}}
    \right),
\end{equation}
where $\hat{u}_{n}$ is the normalized prediction. The final physical prediction
is obtained by de-normalization:
\begin{equation}
    \hat{u}
    =
    \hat{u}_{n}\sigma_y+\mu_y .
\end{equation}
For uncertainty-aware operators such as Bayesian DeepONet, the model additionally
returns a predictive variance, yielding
$\hat{u}(z)\sim\mathcal{N}(\mu_{\phi}(z),\sigma_{\phi}^{2}(z))$.


\subsection{Input Handling Framework}
\label{sec:input_handling_framework}

\begin{figure*}[htbp]
    \centering
    \includegraphics[width=\textwidth]{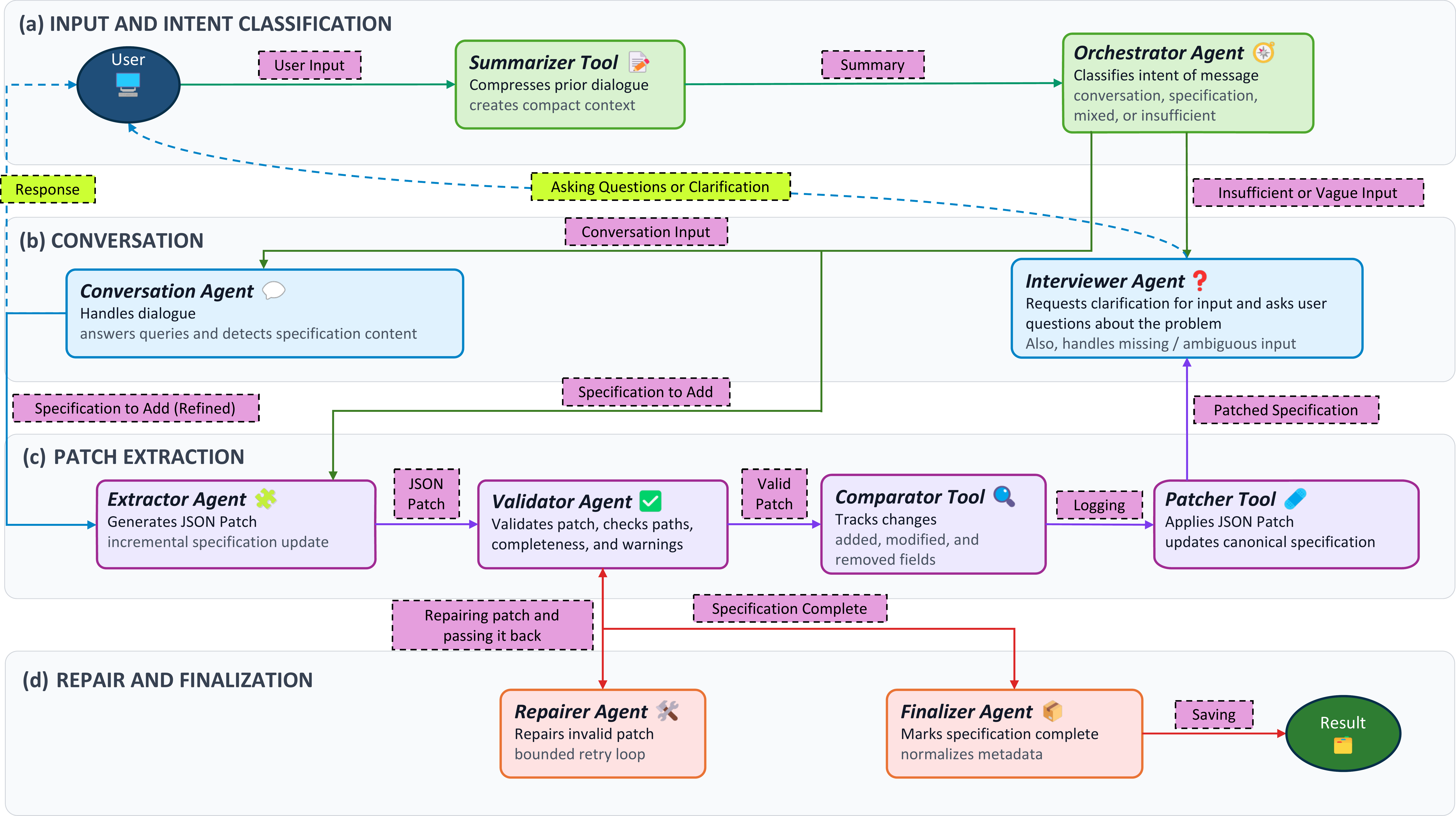}
    \caption{Input-handling agent workflow. Input and Intent Classification (a) summarises the user input, classifies the message, and routes it to the right path. Conversation (b) handles dialogue and clarification without changing the canonical specification. Patch Extraction (c) converts specification content into JSON patches, checks the proposed updates, and applies only valid patches. Repair and Finalization (d) fixes failed patches when possible and marks complete specifications as ready for downstream data generation, solver execution, and operator training.}
    \label{fig:input_handling_agent_system_architecture}
\end{figure*}

\subsubsection{Input and Intent Classification}
\label{sec:input_and_intent_classification}

The input-handling framework converts a user's natural-language request into an executable ODE/PDE specification. Figure~\ref{fig:input_handling_agent_system_architecture} shows the agent graph used for this process; all LLM-based agents in this graph use gpt-o4-mini. A user turn is first classified by the Orchestrator and, when needed, split into conversational and specification-bearing parts. Only the specification-bearing content is allowed to update the canonical problem specification. This prevents ordinary questions, such as "what happens near the centre of the mesh?", from accidentally changing the PDE state.

Each turn is treated as an incremental edit. The graph state stores the latest user message, recent turns, a compressed memory summary, the canonical JSON specification, the candidate patch, a correction log and validation feedback. This stateful design matches how scientific problem entry usually happens: users often provide the equation first, then add domains, coefficients, boundary conditions or solver settings across later turns.

The Orchestrator assigns each message to one of five intent classes: conversation, specification, mixed input, assistant-generated specification or no usable input.

\subsubsection{Conversation}
\label{sec:input_handling_conversation}

Conversation-only turns are answered without editing the canonical specification. Specification turns are routed to the Extractor. Mixed turns are split so that the dialogue part is answered while the executable part proceeds to patch extraction. Assistant-generated specification covers cases where the user gives an informal scientific request and asks the system to turn it into supported mathematical content. Empty or unclear inputs are routed to the Interviewer, which asks for the missing scientific detail rather than guessing.

\subsubsection{Patch Extraction}
\label{sec:patch_extraction}

The Extractor emits JSON patch operations over the current canonical specification; it does not regenerate the full schema. For example, if the user changes only the right boundary from Dirichlet to Neumann, the expected output is a local replacement on the right-boundary field rather than a new equation, mesh, initial condition and solver block. The Comparator records which fields are added, modified or removed, and the Validator checks path validity, operation type, schema consistency and whether the resulting ODE/PDE has the information required for execution.

\subsubsection{Repair and Finalization}
\label{sec:repair_and_finalization}

Invalid patches are not applied. The Repairer receives the current specification, the failed patch and the Validator feedback, then returns a corrected patch for another validation pass. The current implementation allows two repair attempts. If both attempts fail, the Interviewer asks the user for clarification and no partial update is committed. When validation succeeds, the Patcher updates the canonical specification. If the Validator marks the problem complete, the Finalizer sets the problem status, schema version and metadata required by downstream dataset generation, solver execution and operator training. Graph-state fields, agent/tool responsibilities, completion checks and validation scenarios are listed in Appendix~\ref{app:input_handling}.


\begin{figure*}[htbp]
    \centering
    \includegraphics[width=\textwidth]{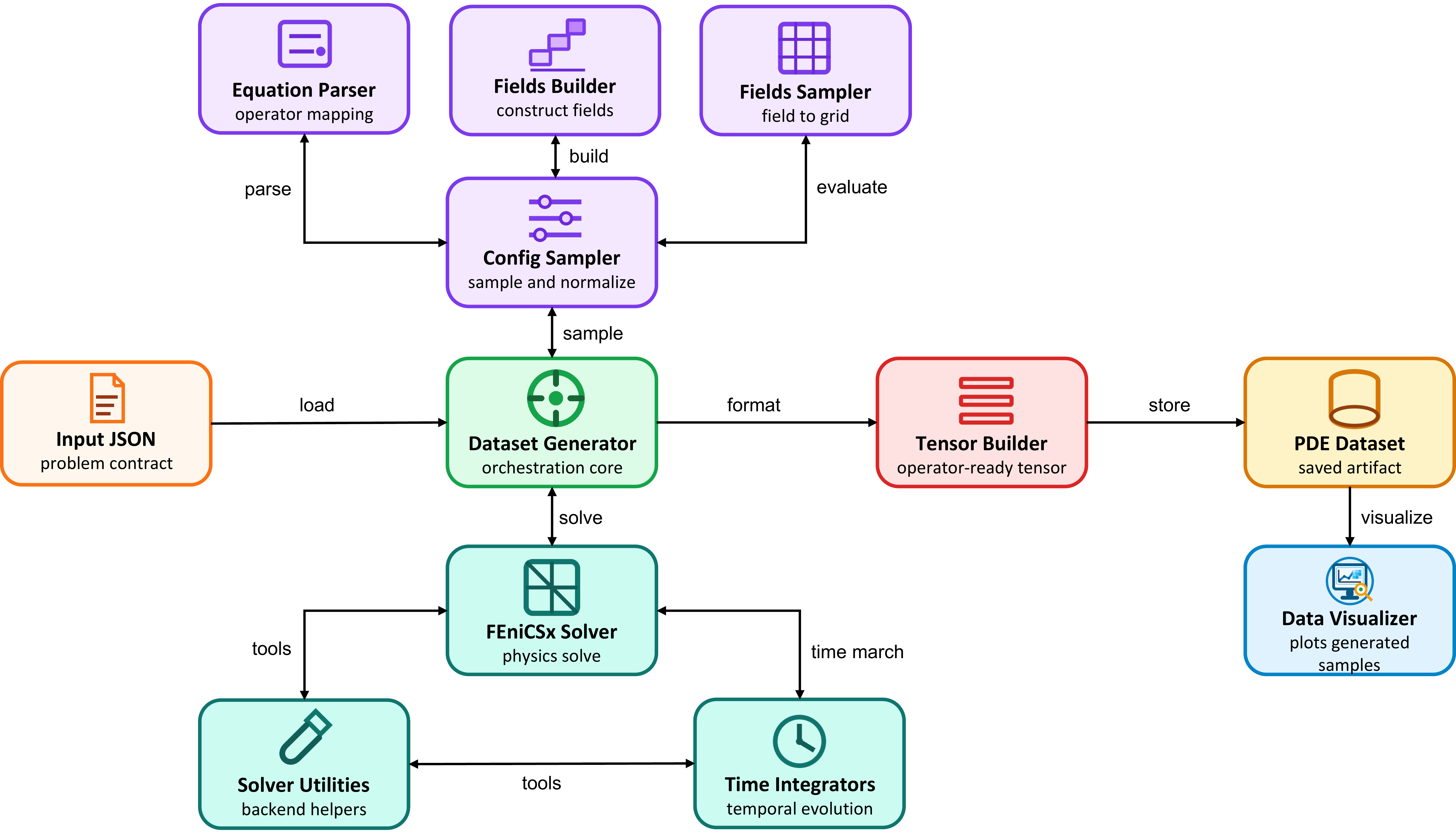}
    \caption{Data-generation workflow for converting an input JSON problem specification into a solver-backed operator dataset. The module parses the governing equation, samples scalar and field-valued parameters, evaluates fields on the requested grids, solves the configured ODE/PDE using the FEniCSx backend and time integrators, and formats valid solutions into operator-ready tensors for storage and visualisation.}
    \label{fig:data_generation_system_architecture}
\end{figure*}

\subsection{Data Generation Framework}
\label{sec:data_generation_framework}

The data-generation framework converts a validated problem specification into solver-generated datasets for operator learning. As shown in Figure~\ref{fig:data_generation_system_architecture}, the workflow starts from an input JSON file, parses the governing equation, samples scalar and field-valued quantities, solves the configured ODE/PDE problem, and stores the result as an operator-ready dataset artifact.

\subsubsection{Configuration Parsing and Sampling}

The system receives a JSON specification containing the domain, optional time range, governing equation, coefficient definitions, initial condition, boundary conditions, and data-generation settings. The equation parser maps symbolic terms such as $u_t$, $u_{xx}$, $u_x$, and $u$ to solver coefficients. The config sampler reads the JSON specification, samples scalar parameters, normalises boundary-condition entries, and infers whether the problem is steady or time-dependent. Field specifications are constructed and evaluated on the requested one- or two-dimensional grids so that coefficient fields, initial conditions, and boundary-condition fields can be passed to the FEniCSx solver and stored as conditioning inputs for downstream operator training. Some notable features of the data-generation framework are listed in Appendix~\ref{sec:supported_data_generation_features}.

\subsubsection{Solver Integration and Dataset Generation}

For each sample, the config sampler produces a solver configuration containing coefficient values, callable fields, initial conditions, boundary-condition types, and boundary parameters. The FEniCSx solver returns either a steady solution array or a time-indexed trajectory. The solver supports edge-wise Dirichlet, Neumann, and Robin boundary conditions. For time-dependent cases, it provides seven time-integration schemes, which are described in Appendix~\ref{sec:supported_data_generation_features}. Further details on the finite-element weak form, solver workflow, and validation studies are given in Appendix~\ref{sec:fenicsx_solver_backend}.

The dataset generator checks each computed solution for numerical validity and retains only samples that pass this check. Accepted samples are collected and passed to the tensor builder. The saved dataset has the form
\begin{equation}
    \mathcal{D}
    =
    \left\{
        X_f,\; X_s,\; Z,\; Y,\; \mathcal{M}
    \right\},
\end{equation}
where $X_f$ denotes conditioning fields, $X_s$ denotes conditioning scalars, $Z$ denotes query coordinates, $Y$ denotes solver targets, and $\mathcal{M}$ denotes metadata. In the current implementation, these entries are stored as branch fields, branch scalars, trunk coordinates, targets, and metadata. The metadata preserves coordinate ordering, tensor layout, conditioning-variable names, spatial axes, time values, and output-scaling statistics used during training and inference.


\subsection{Operator Training Module}
\label{sec:operator_training_module}

The operator training module trains a selected neural operator on the dataset produced by the data-generation stage. The dataset loader reads conditioning fields, conditioning scalars, query coordinates, targets, and metadata, then prepares mini-batches containing the inputs and solver-generated target values required by the selected operator.

\begin{figure*}[htbp]
    \centering
    \includegraphics[width=\textwidth]{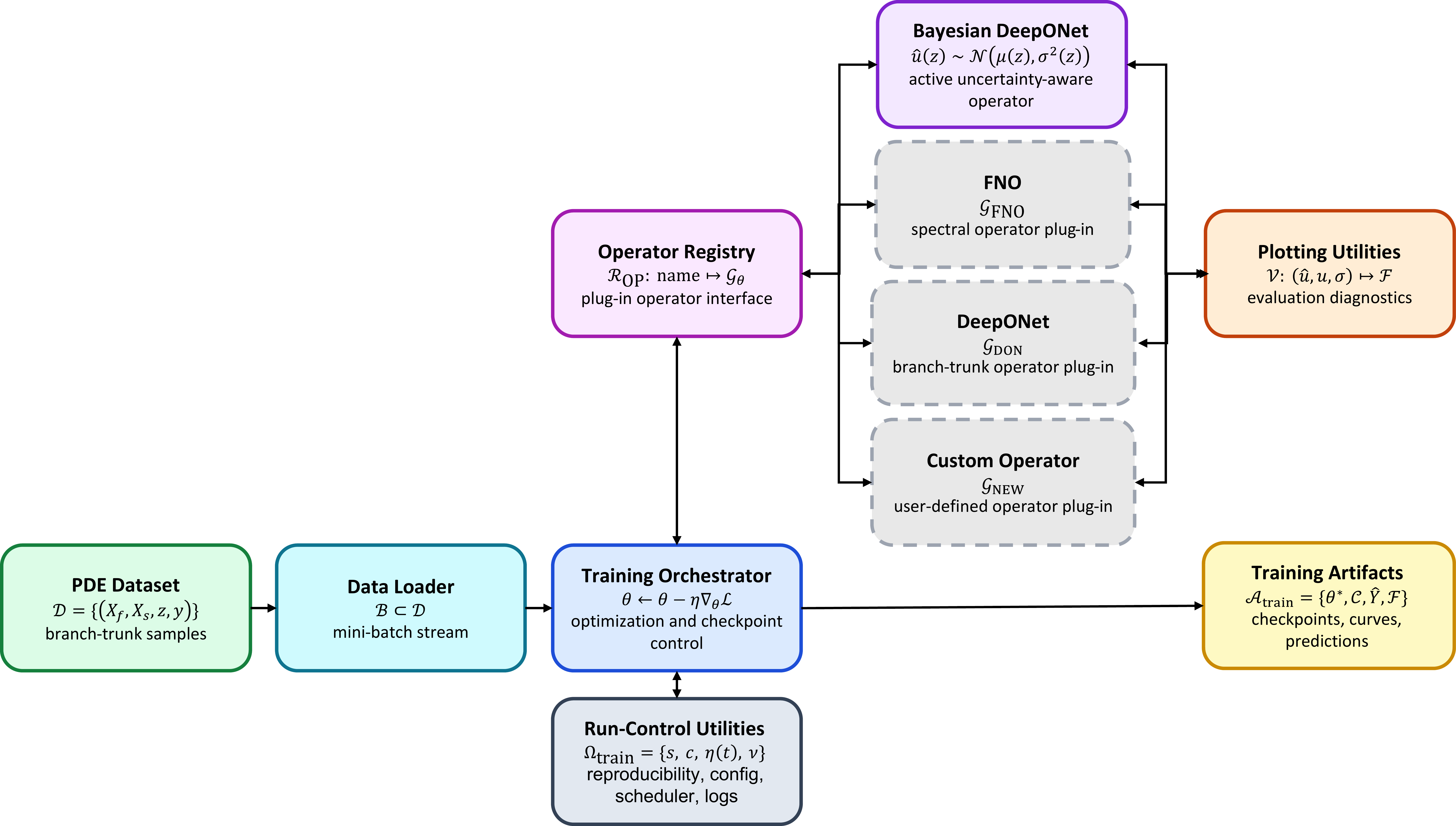}
    \caption{Operator training workflow with a registry-based interface for neural-operator selection. The module loads operator-ready datasets, prepares mini-batches, dispatches training to the selected operator, saves checkpoints and training artifacts, and generates evaluation diagnostics through shared plotting utilities.}
    \label{fig:operator_training_system_architecture}
\end{figure*}

As shown in Figure~\ref{fig:operator_training_system_architecture}, the training orchestrator manages run-level tasks such as dataset loading, tensor checks, device selection, hyperparameter loading, checkpointing, and logging. Model-specific operations are delegated to the selected operator through the operator registry. This keeps the training entry point independent of the operator class, allowing Bayesian DeepONet, FNO, deterministic DeepONet, or user-defined operators to be registered under the same interface.

In the current implementation, Bayesian DeepONet is the active operator. The saved dataset uses DeepONet-compatible entries for field inputs, scalar inputs, query coordinates, targets, and metadata. Bayesian DeepONet maps these entries through field and scalar branches, a trunk network, and stochastic Bayesian layers for uncertainty-aware prediction. During training, the loop records loss curves and saves the best checkpoint. During evaluation, the best checkpoint is restored, predictions are mapped back to physical units, and plotting utilities generate diagnostics such as predicted fields, reference fields, uncertainty estimates, and training curves. A broader description of the multi-branch Bayesian DeepONet is provided in Appendix~\ref{sec:multi_branch_bayesian_deeponet}.


\subsection{Inference Module}
\label{sec:inference_module}

The inference module generates predictions from a new JSON specification using a trained neural operator. It does not call the numerical solver. Instead, it rebuilds the required conditioning inputs and query coordinates from the inference specification, restores the trained checkpoint, runs the operator forward pass, and saves the resulting predictions.

\begin{figure*}[htbp]
    \centering
    \includegraphics[width=\textwidth]{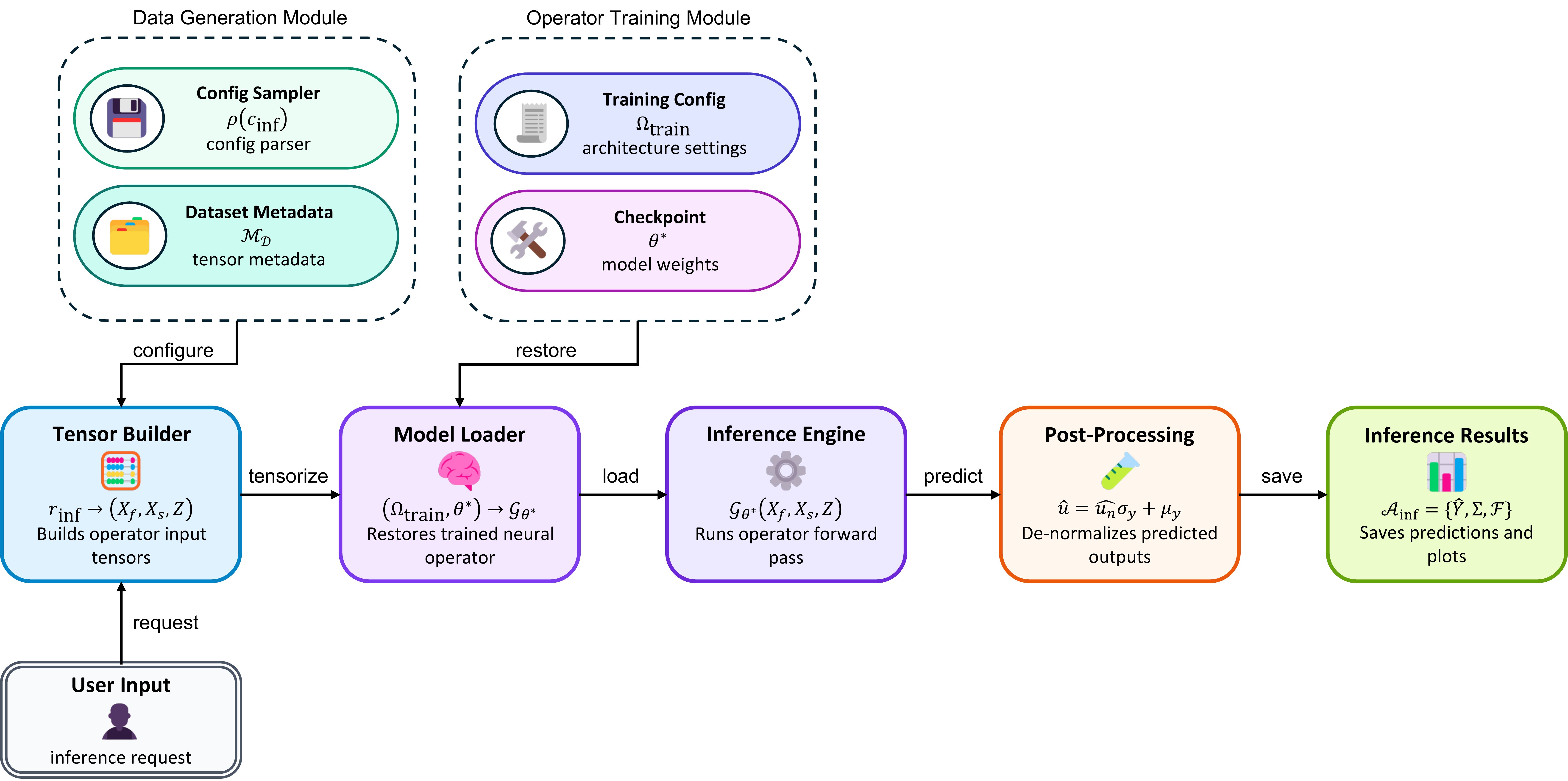}
    \caption{Inference workflow for generating solver-free neural-operator predictions from a new problem specification. The module rebuilds inference tensors using the saved dataset metadata, restores the trained checkpoint, runs the operator forward pass, de-normalises the outputs, and saves prediction plots and numerical inference results.}
    \label{fig:inference_system_architecture}
\end{figure*}

As shown in Figure~\ref{fig:inference_system_architecture}, the workflow starts from a user-provided inference request. The config sampler parses the new specification, while the dataset metadata provides the field ordering, scalar ordering, coordinate layout, spatial axes, time values, and output-scaling statistics used during training. The tensor builder then constructs the conditioning fields, conditioning scalars, and query coordinates for the new problem instance.

The model loader restores the trained operator using the training configuration, dataset metadata, and saved checkpoint. The inference engine evaluates the model over all query coordinates. In the current Bayesian DeepONet implementation, multiple stochastic forward passes are used to estimate the predictive mean and predictive standard deviation.

The post-processing stage converts normalised predictions back to physical units and saves the inference outputs. The saved artefacts include problem-specific plots and a numerical result file containing predictions, uncertainty estimates, query coordinates, grids, sampled parameters, and the specification name.


\section{Experiments and Results}
\label{sec:results}


\subsection{Agentic Workflow Accuracy}
\label{sec:agentic_workflow_accuracy}

We evaluate the input-handling workflow on 70 scripted ODE/PDE scenarios from the validation runners: 15 ODE, 30 PDE (1D) and 25 PDE (2D) cases. Each scenario is processed by the agent graph, and the generated JSON specification is compared with a reference specification. The scenarios include missing coefficients, unsupported equations, sampled parameters, boundary-condition edits and invalid inputs that are corrected in later turns. Representative scenarios are listed in Appendix~\ref{app:validation_scenarios}.

We compare the complete system with three variants. The first removes the Repairer, so invalid patches are not corrected. The second removes the Validator, so patches are applied without schema and completion checks. The third is a single-shot baseline that maps the full dialogue directly to a JSON specification.

Table~\ref{tab:overall_model_accuracy} shows that the complete system achieves the highest aggregate accuracy at 81.43\%. Removing the Validator gives the largest drop, reducing accuracy to 54.29\%. This indicates that validation is needed to catch incorrect paths, missing fields and schema errors before they enter the specification. Removing the Repairer also lowers accuracy to 65.71\%, showing that many failed patches are close to correct but need small fixes before application.

\begin{table}[htbp]
    \centering
    \caption{Accuracy across difficulty levels for the complete system and ablated variants.}
    \label{tab:overall_model_accuracy}
    \begin{tabular}{lc}
        \toprule
        \textbf{Model} & \textbf{Accuracy (\%)} \\
        \midrule
        \textbf{Complete System} & \textbf{81.43} \\
        No Repairer & 65.71 \\
        No Validator & 54.29 \\
        Single Shot & 74.29 \\
        \bottomrule
    \end{tabular}
\end{table}

\begin{figure}[htbp]
    \centering
    \includegraphics[width=0.8\linewidth]{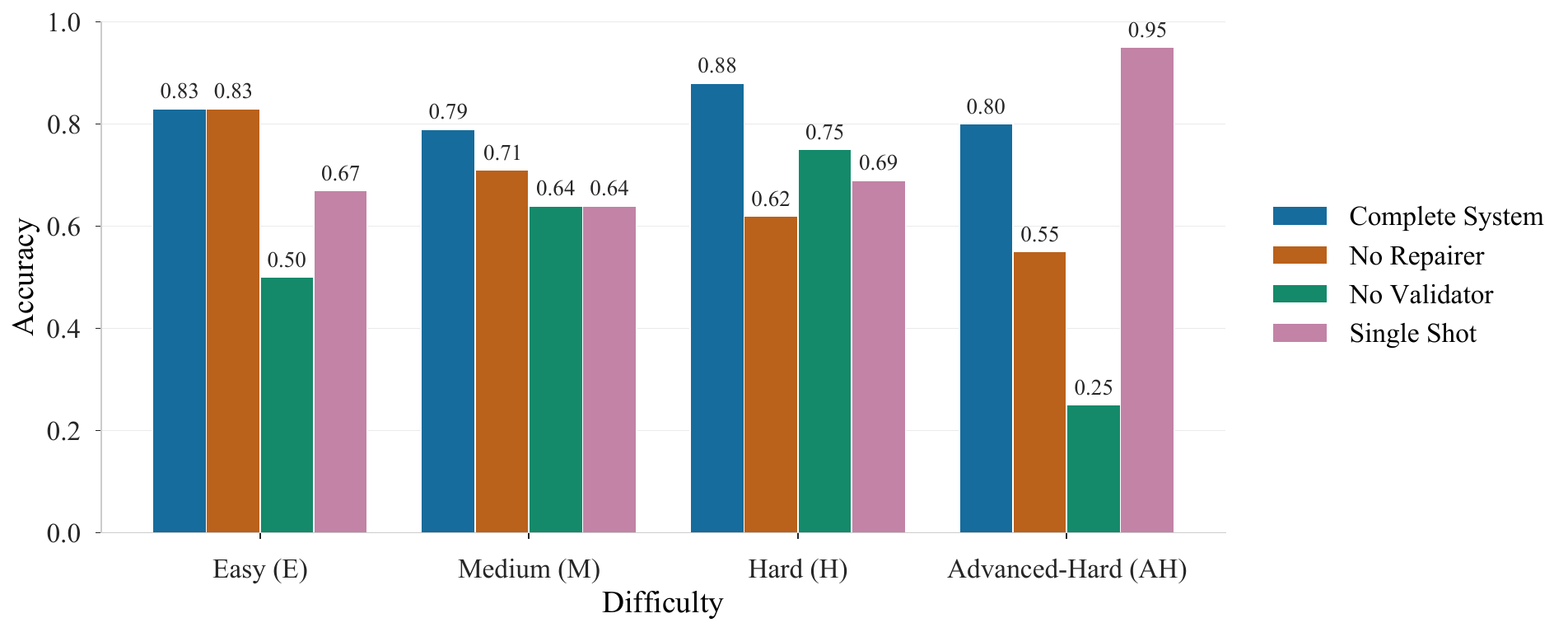}
    \caption{Aggregate accuracy across scripted ODE and PDE specification scenarios.}
    \label{fig:success_rate_1}
\end{figure}

The accuracy by difficulty level is reported in Figure~\ref{fig:success_rate_1}. The complete system performs well on easy, medium and hard cases, with accuracies of 0.83, 0.79 and 0.88. The no-Validator variant performs poorly on advanced-hard cases, dropping to 0.25, because invalid intermediate updates can corrupt the state. The single-shot baseline performs best on advanced-hard cases, reaching 0.95, since it sees the full corrected dialogue at once rather than maintaining a valid state after every turn.

\begin{figure}[htbp]
    \centering
    \includegraphics[width=0.8\linewidth]{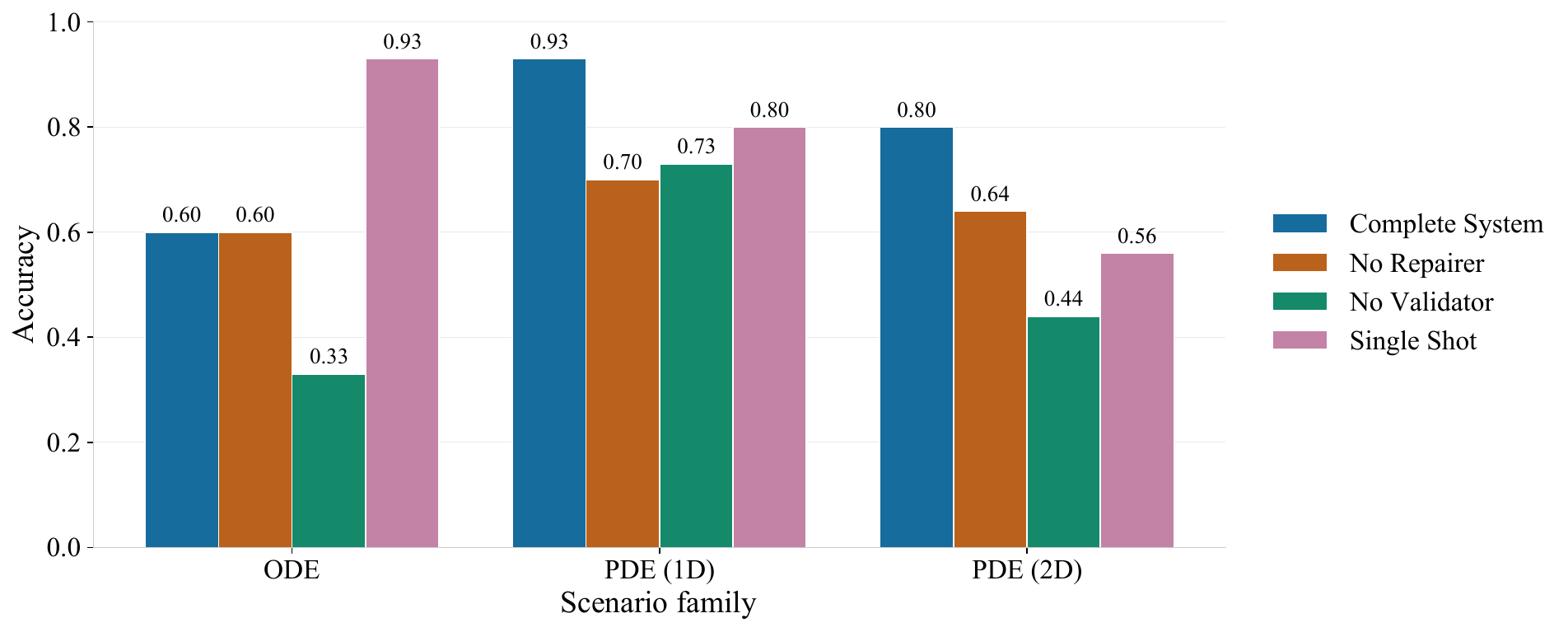}
    \caption{Accuracy across ODE, PDE (1D) and PDE (2D) scenario families.}
    \label{fig:success_rate_2}
\end{figure}

The accuracy by equation family is grouped in Figure~\ref{fig:success_rate_2}. The complete system performs best on PDE cases, reaching 0.93 on PDE (1D) and 0.80 on PDE (2D). These cases benefit from validation and repair because the specifications contain linked fields such as domains, grids, coefficients and boundary conditions. ODE cases are shorter and closer to complete specifications, so the single-shot baseline performs better on that family. Overall, the results show that validation and repair are most useful when the specification is structured and updated over multiple turns.


\subsection{Neural Operator Evaluation}
\label{sec:neural_operator_evaluation}

We evaluate the trained neural operator against solver-generated reference solutions on benchmark ODE/PDE tasks. Detailed prediction plots, uncertainty maps, loss histories and training summaries are provided in Appendix~\ref{sec:training_results}, while runtime comparisons are reported separately in Appendix~\ref{sec:runtime_analysis}. Tables~\ref{tab:operator_training_results} and~\ref{tab:operator_training_results_continued} report the training metrics, and Table~\ref{tab:runtime_inference_speedup} reports per-sample data-generation time, neural-operator inference time and speedup. The benchmark set includes temporal ODEs, steady one-dimensional problems, transient one-dimensional PDEs, steady two-dimensional PDEs and transient two-dimensional PDEs. All tasks use the same dataset interface, with conditioning fields, conditioning scalars, query coordinates and normalised targets passed to the operator trainer.

The training summary reports the best training epoch, best training loss, MSE at the saved best checkpoint and MSE at the final checkpoint. The best checkpoint is selected using the lowest average training loss, since the final epoch is not always the best-performing epoch, especially for transient or higher-dimensional problems.

\begin{figure}[htbp]
    \centering
    \includegraphics[width=0.8\linewidth]{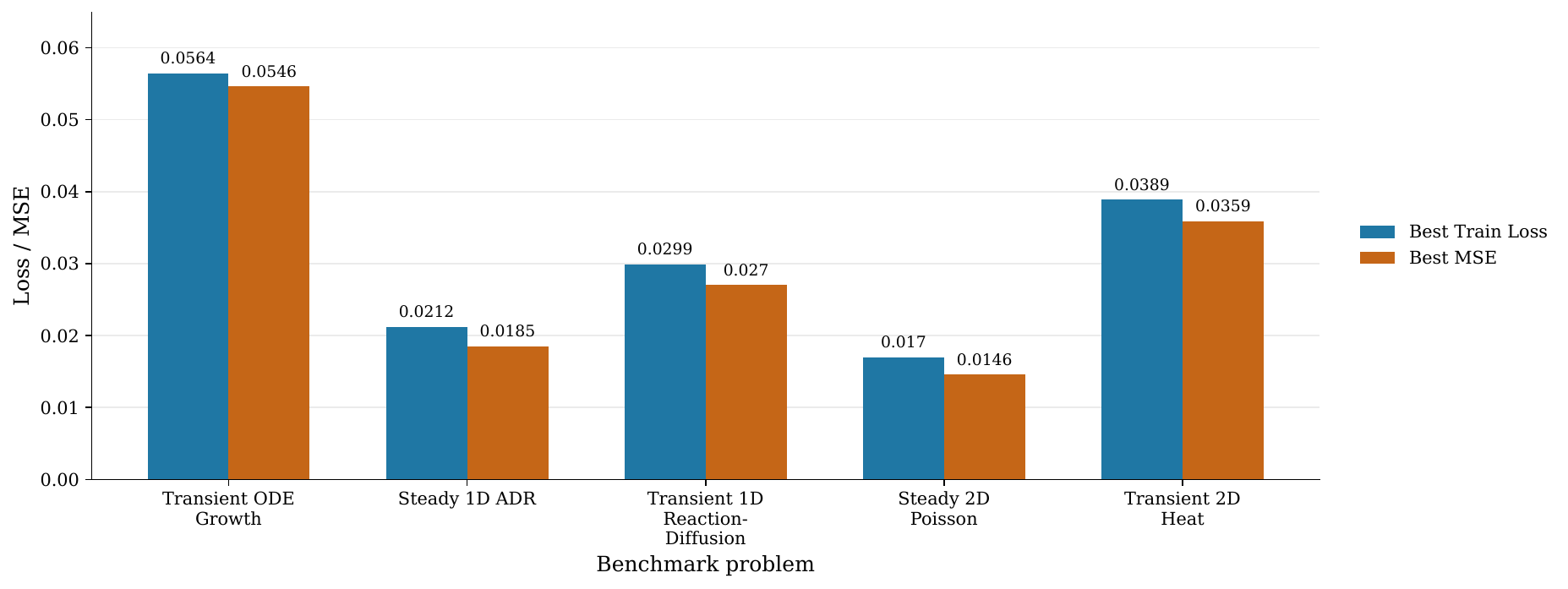}
    \caption{Strong-performing representative benchmarks selected from the main problem classes, comparing best training loss and best-checkpoint MSE.}
    \label{fig:loss_mse_bargraph}
\end{figure}

Most steady and lower-dimensional cases obtain lower errors, while transport-dominated, strongly parametric and higher-dimensional transient cases are harder. The largest errors appear in cases where the solution changes sharply in space or time. Figure~\ref{fig:loss_mse_bargraph} gives a compact visual comparison for representative benchmarks from the main problem classes.

The results indicate that the same training pipeline can handle different equation families without changing the trainer interface. The remaining errors suggest that the harder cases may benefit from larger training sets, denser query grids, improved hyperparameter tuning, longer training runs and operator architectures better matched to the underlying PDE dynamics.

\section{Future Works}
\label{sec:future_work}

The current system uses an operator registry and a shared training interface, but operator selection is still manual. A useful next step is to compare registered operators across the benchmark families used here, including steady elliptic problems, transient diffusion, advection-dominated equations, reaction--diffusion systems and mixed-boundary cases. These comparisons could support an operator-selection agent that chooses a model based on the detected problem class, grid structure, coefficient type, boundary conditions and requested uncertainty output.

The solver backend is currently limited to one- and two-dimensional rectangular domains with linear PDEs. Extending the solver path to irregular geometries would allow the input specification to include geometry, mesh size and boundary tags before simulation. Support for nonlinear PDEs would require nonlinear residual assembly, nonlinear time stepping and additional validation checks.

The platform can also be connected to external preprocessing and postprocessing tools. Gmsh can support geometry and mesh generation, while ParaView can support visual inspection of solver outputs, learned predictions and uncertainty fields. This would move the system toward an agent-controlled workflow for specification, simulation, dataset generation, operator training, inference and visual analysis.

\nocite{*}
\bibliographystyle{plainnat}
\bibliography{references}

\clearpage 

\section{Technical Appendix}
\label{sec:technical_appendix}


\subsection{Supported Data-Generation Features}
\label{sec:supported_data_generation_features}

Table~\ref{tab:data_generation_supported_features} summarises the current scope of the data-generation framework. The framework supports temporal ODEs, spatial ODEs, one- and two-dimensional PDEs, edge-wise boundary conditions, sampled coefficients and multiple time-integration schemes. This table defines the solver-facing feature set used by the benchmark studies in the main paper.

\begin{table}[htbp]
\centering
\caption{Supported capabilities of the data generation framework.}
\label{tab:data_generation_supported_features}
\small
\renewcommand{\arraystretch}{1.35}
\setlength{\tabcolsep}{6pt}

\begin{tabularx}{0.95\linewidth}{|p{0.30\linewidth}|X|}
\hline
\textbf{Capability} & \textbf{Supported Options} \\
\hline

Supported Problem Classes &
Temporal ODE; Spatial ODE; Transient 1D PDE; Steady 2D PDE; Transient 2D PDE. \\
\hline

Supported Boundary Conditions &
Dirichlet, Neumann, and Robin boundary conditions, specified independently on each edge. \\
\hline

Supported Field Types &
Constant Fields, Parametric Scalar Fields, Analytical Fields, and Parametric Analytical Fields. \\
\hline

Supported Distribution Types &
Uniform, Normal, Lognormal, Beta, and Truncated Normal distributions. \\
\hline

Supported Time Schemes &
Backward Euler (BE), Crank--Nicolson (CN), second-order Backward Differentiation Formula (BDF2), second-order Singly Diagonally Implicit Runge--Kutta (SDIRK2), Trapezoidal Rule--Backward Differentiation Formula 2 (TR-BDF2), third-order Strong Stability Preserving Runge--Kutta (SSP-RK3), and Implicit--Explicit Euler (IMEX Euler). \\
\hline

\end{tabularx}
\end{table}

\subsection{Multi-Branch Bayesian DeepONet Architecture}
\label{sec:multi_branch_bayesian_deeponet}

\begin{figure*}[htbp]
    \centering
    \includegraphics[width=\textwidth]{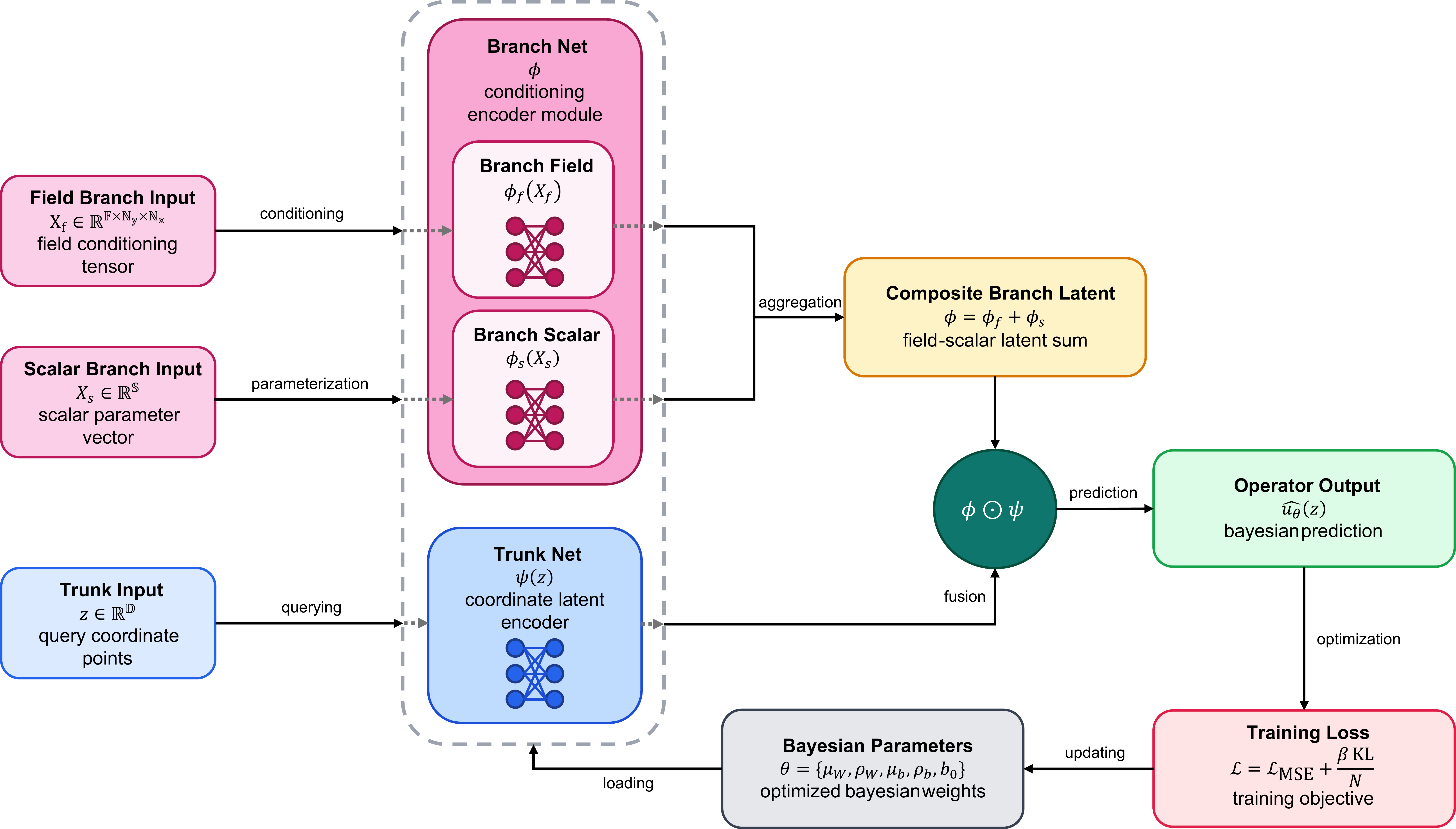}
    \caption{Multi-branch Bayesian DeepONet architecture for mapping conditioning inputs and query coordinates to solution values. Field-valued inputs and scalar parameters are encoded by separate branch networks, query coordinates are encoded by the trunk network, and the fused latent representation produces a Bayesian prediction trained with an MSE--KL objective.}
    \label{fig:bayesian_deeponet_system_architecture}
\end{figure*}

The operator module uses a multi-branch Bayesian DeepONet to map simulation conditions and query coordinates to solution values. As shown in Figure~\ref{fig:bayesian_deeponet_system_architecture}, the model receives field-valued conditioning tensors $X_f$, scalar conditioning vectors $X_s$, and query points $z$. Field inputs encode spatially varying quantities, scalar inputs encode global parameters, and the trunk network evaluates the requested solution location.

Missing input types are handled explicitly. If a dataset has no field-valued or scalar inputs, the corresponding branch is bypassed and replaced with a zero latent. This keeps the forward pass fixed across temporal ODEs, steady PDEs, and transient PDEs. Each active branch is implemented as a Bayesian multilayer perceptron with GELU activations, layer normalisation, and dropout. The trunk uses the same style of network for query coordinates, and all branches are projected into a shared latent dimension before fusion.

For a single query,
\begin{equation}
    \phi_f = B_f(X_f), \qquad
    \phi_s = B_s(X_s), \qquad
    \psi = T(z),
\end{equation}
with the conditioning representation
\begin{equation}
    \phi = \phi_f + \phi_s .
\end{equation}
The DeepONet interaction is then
\begin{equation}
    h(z)=\phi\odot\psi(z),
    \qquad
    \hat{u}_n(z)=Wh(z)+b .
\end{equation}
The network predicts normalised targets; physical units are recovered outside the model using dataset statistics stored during training.

Uncertainty is introduced through Bayesian linear layers. Each weight is sampled as
\begin{equation}
    W = \mu_W + \sigma_W \epsilon,
    \qquad
    \epsilon \sim \mathcal{N}(0,I),
\end{equation}
where $\sigma_W$ is obtained through a softplus transform. Training minimises
\begin{equation}
    \mathcal{L}
    =
    \mathcal{L}_{\mathrm{MSE}}
    +
    \lambda_{\mathrm{KL}}
    \frac{\mathrm{KL}}{N_{\mathrm{train}}},
\end{equation}
with $\lambda_{\mathrm{KL}}=10^{-4}$. During evaluation, repeated stochastic forward passes produce the predictive mean and empirical variance, giving both operator predictions and uncertainty estimates from the same model.

This design keeps the current implementation tied to Bayesian DeepONet while leaving the surrounding pipeline less dependent on a single operator class. The dataset, trainer, checkpoint loader, and inference code pass conditioning inputs and query coordinates through a common interface; only the registered operator decides how those tensors are fused internally. In the present experiments, the branch--trunk product is used for DeepONet, but the same outer workflow can register a different neural operator without changing the data-generation or inference stages.


\subsection{Training Results}
\label{sec:training_results}

All experiments used a fixed configuration with 1050 total samples, split into 1000 training samples and 50 test samples. Each model was trained for 150 epochs with $\lambda_{\mathrm{KL}}=10^{-4}$. During evaluation, 10 test samples were used for case-level plots, and Bayesian uncertainty was estimated using 100 Monte Carlo forward passes per evaluated sample.

Tables~\ref{tab:operator_training_results} and~\ref{tab:operator_training_results_continued} summarise the benchmark results across temporal ODEs, steady 1D and 2D problems, and transient PDEs. The best-checkpoint MSE values are generally close to the final-epoch MSE values, so late-epoch drift is limited in most cases.

Figures~\ref{fig:ode_pure_temporal_reaction}--\ref{fig:transient_2d_heat_gaussian_jet} show representative predictions, solver references, uncertainty estimates, and loss histories. The MSE term accounts for most of the reduction in the objective, while the KL term remains active as Bayesian regularisation.

Accuracy depends on the physics of the problem. Smooth steady problems and lower-dimensional systems are fitted closely, especially when the solution map changes smoothly with the sampled parameters. This is visible in the steady 1D and elliptic 2D cases, where the predicted fields remain close to the solver references. Transport-dominated and strongly parameterised transient cases are harder, with larger errors near sharp gradients, moving features or localised source effects. The uncertainty maps usually highlight these regions, giving a useful qualitative diagnostic alongside the prediction error and pointing to the need for operator choices or coordinate encodings that better represent moving and localised structures.

\clearpage
\subsubsection*{\underline{\textbf{PROBLEM 1: Forced Temporal Reaction ODE}}}

\begin{figure}[h]
    \centering
    \includegraphics[width=\textwidth]{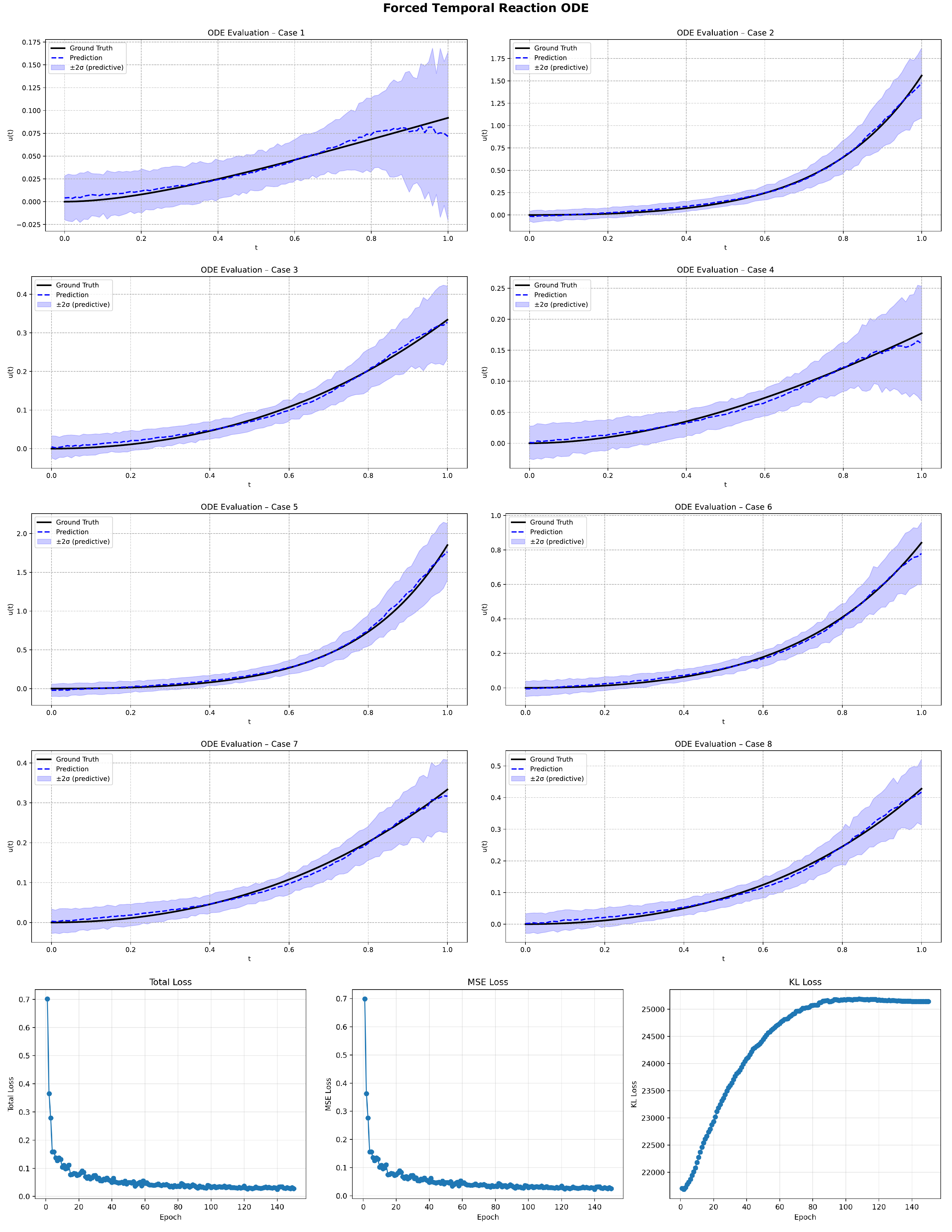}
    \caption{Forced temporal reaction ODE evaluation with Bayesian DeepONet predictions, predictive uncertainty bands, and training loss histories.}
    \label{fig:ode_pure_temporal_reaction}
\end{figure}

For the forced temporal reaction ODE, the solution evolves only in time under a reaction term and external forcing. Figure~\ref{fig:ode_pure_temporal_reaction} shows that the Bayesian DeepONet captures both slow and faster-growing trajectories across the evaluated cases. The predictions follow the solver-generated curves closely, with larger uncertainty mainly visible for higher-amplitude responses. This suggests that the model learns the temporal response well while still reflecting uncertainty where the dynamics grow more strongly. 

\FloatBarrier
\clearpage

\subsubsection*{\underline{\textbf{PROBLEM 2: Steady 1D Variable Coefficient Advection-Diffusion-Reaction Problem}}}
\begin{figure*}[h]
    \centering
    \includegraphics[width=\textwidth]{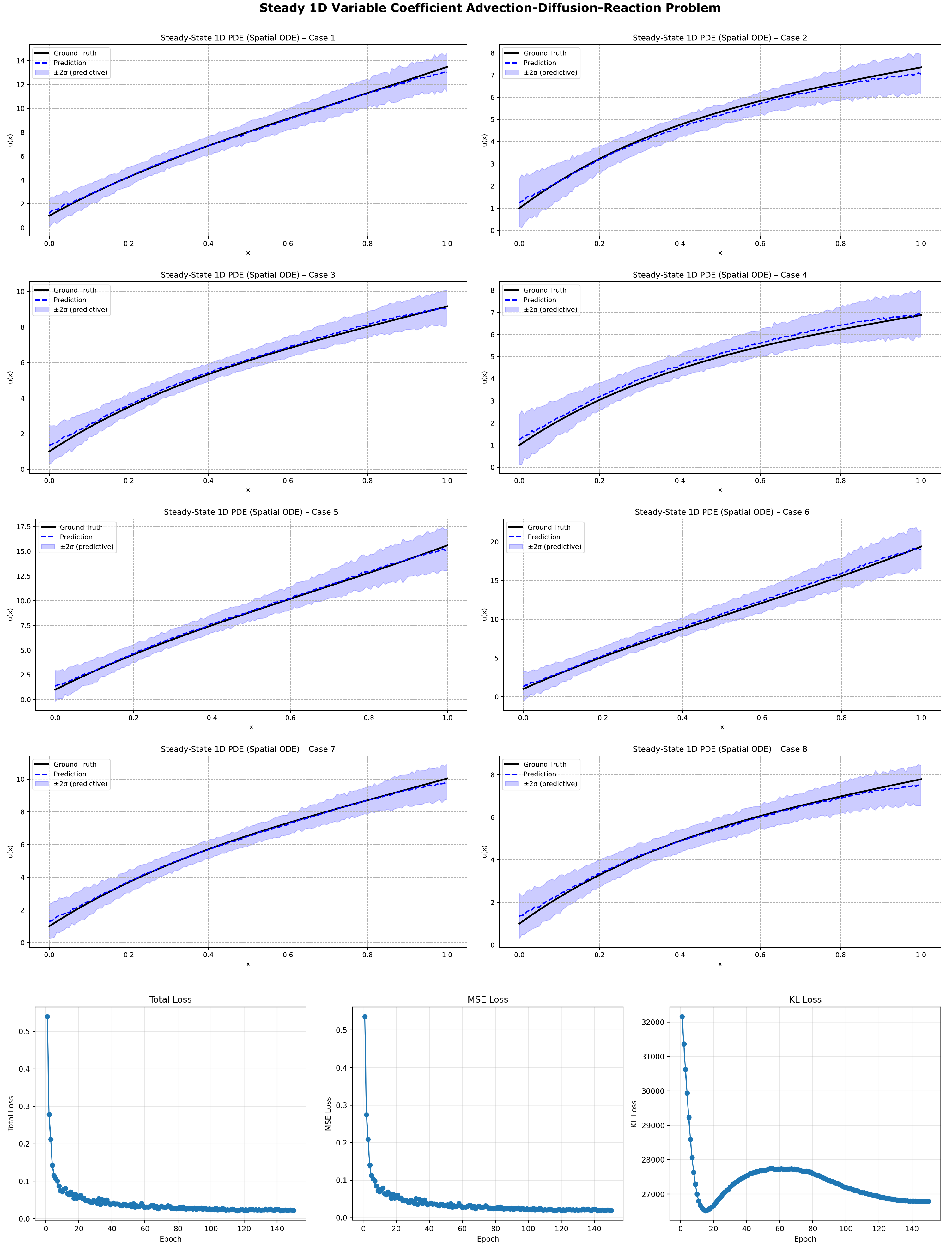}
    \caption{Steady 1D variable-coefficient advection--diffusion--reaction evaluation with Bayesian DeepONet predictions and training loss histories.}
    \label{fig:steady_1d_adf}
\end{figure*}

For the steady 1D advection--diffusion--reaction problem, the solution is governed by diffusion, transport and reaction effects along the spatial domain. Figure~\ref{fig:steady_1d_adf} shows that the Bayesian DeepONet follows the solver-generated profiles across coefficient samples, with small errors mainly where the solution rises more sharply. The narrow predictive bands indicate stable uncertainty estimates for this steady 1D case.

\FloatBarrier

\clearpage
\subsubsection*{\underline{\textbf{PROBLEM 3: Steady 1D Parametric Helmholtz Problem}}}
\begin{figure*}[h]
    \centering
    \includegraphics[width=\textwidth]{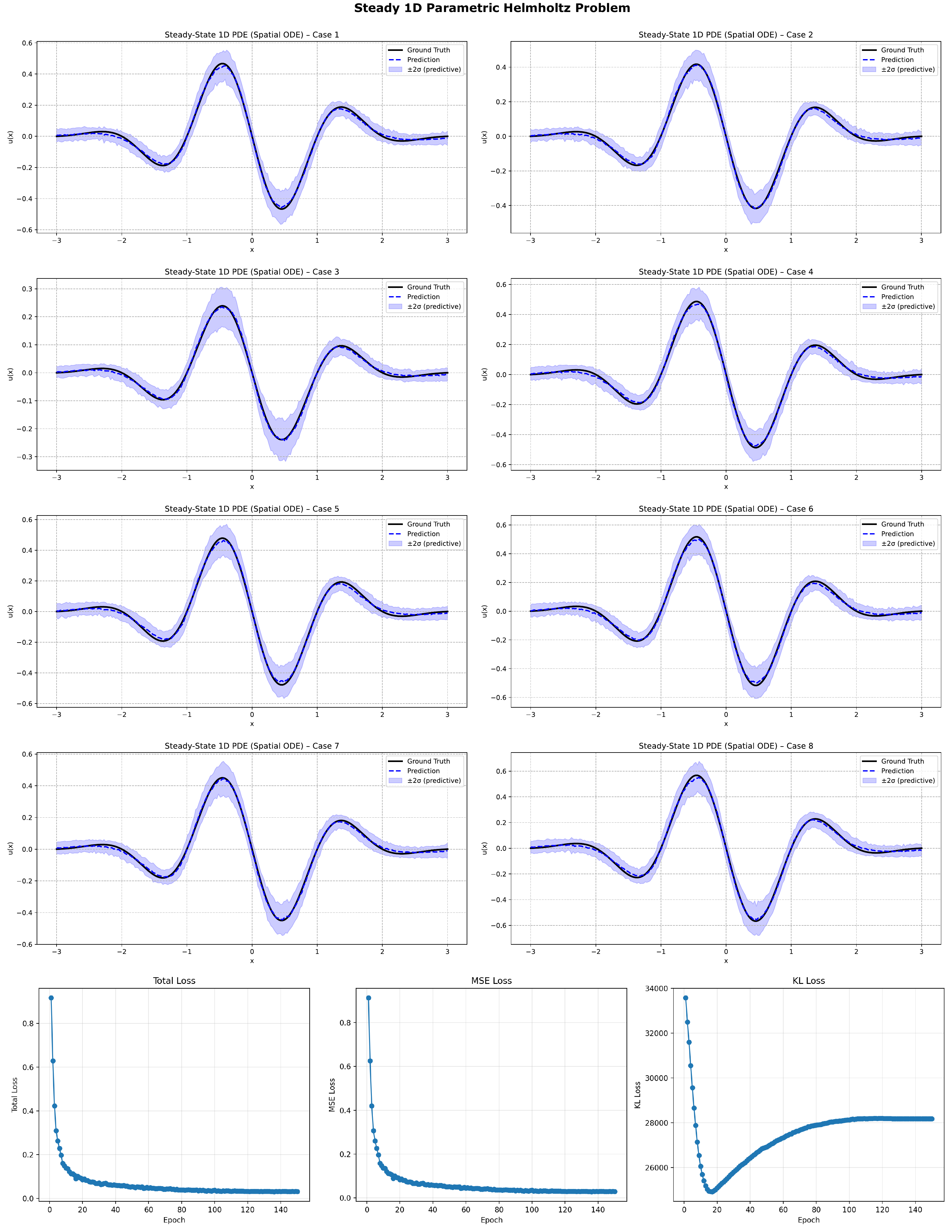}
    \caption{Bayesian DeepONet evaluation for the steady 1D parametric Helmholtz problem.}
    \label{fig:steady_1d_helmholtz}
\end{figure*}

For the steady 1D parametric Helmholtz problem, the solution is governed by an oscillatory response induced by the Helmholtz coefficient and the spatial forcing term. Figure~\ref{fig:steady_1d_helmholtz} shows that the Bayesian DeepONet captures the main shape and sign changes of the solver-generated profiles across parameter samples. The predictions remain close to the reference curves over most of the domain, indicating that the model learns the parametric dependence of this steady oscillatory problem. 

\FloatBarrier

\clearpage
\subsubsection*{\underline{\textbf{PROBLEM 4: Steady 2D Anisotropic Cross Diffusion Problem}}}

\begin{figure*}[h]
    \centering
    \includegraphics[width=0.9\textwidth]{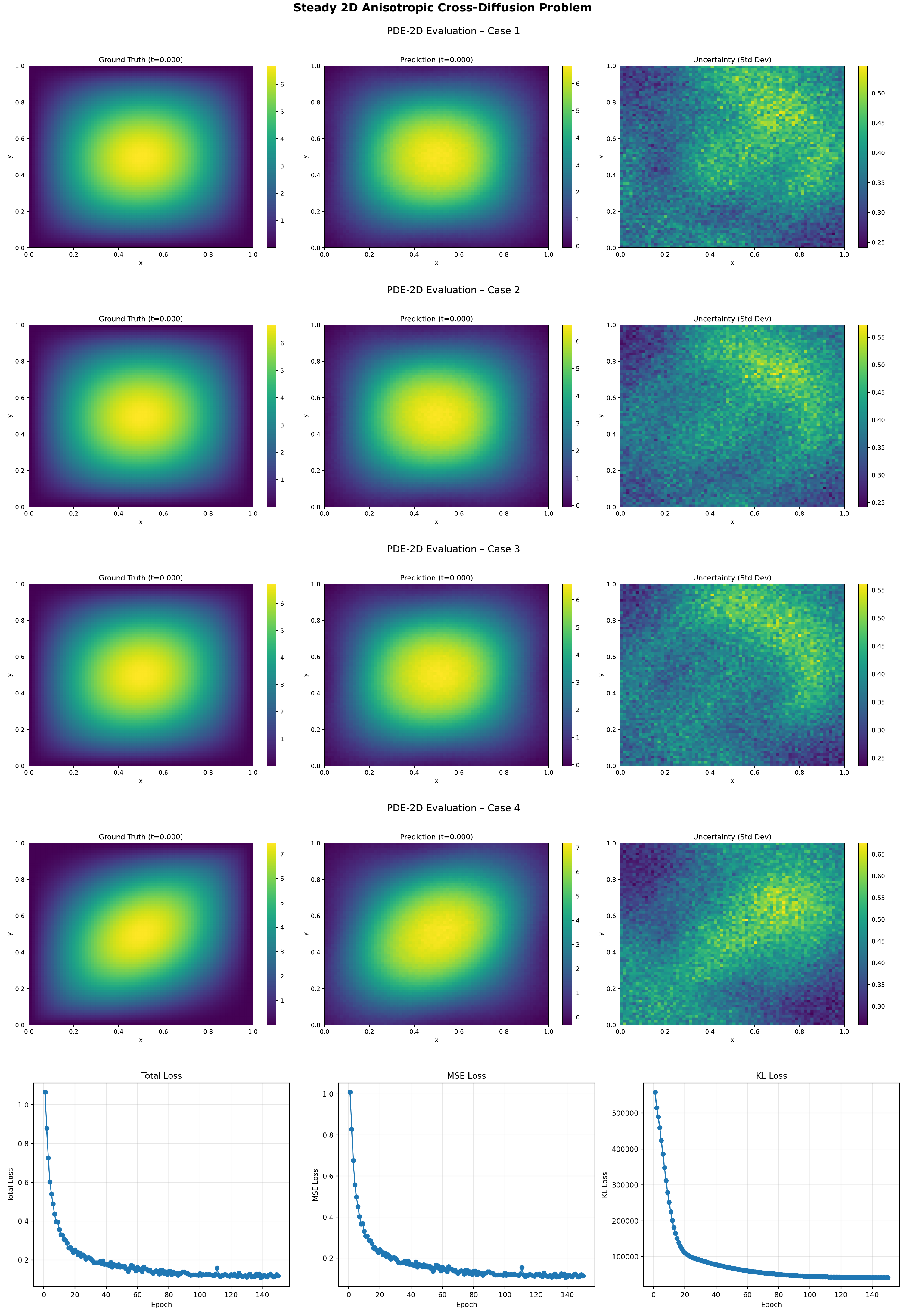}
    \caption{Bayesian DeepONet evaluation for the steady 2D anisotropic cross-diffusion problem.}
    \label{fig:steady_2d_cross_term}
\end{figure*}

For the steady 2D cross-diffusion problem, the solution depends on coupled diffusion across both spatial directions, including mixed-derivative effects from the cross terms. Figure~\ref{fig:steady_2d_cross_term} shows that the Bayesian DeepONet captures the main field structure across the evaluated coefficient samples. The predicted fields are close to the solver-generated references, with uncertainty concentrated in regions where the coupled diffusion response changes more strongly.

\FloatBarrier
\clearpage

\subsubsection*{\underline{\textbf{PROBLEM 5: Steady 2D Poisson Problem with Mixed Boundary Conditions}}}

\begin{figure*}[h]
    \centering
    \includegraphics[width=0.9\textwidth]{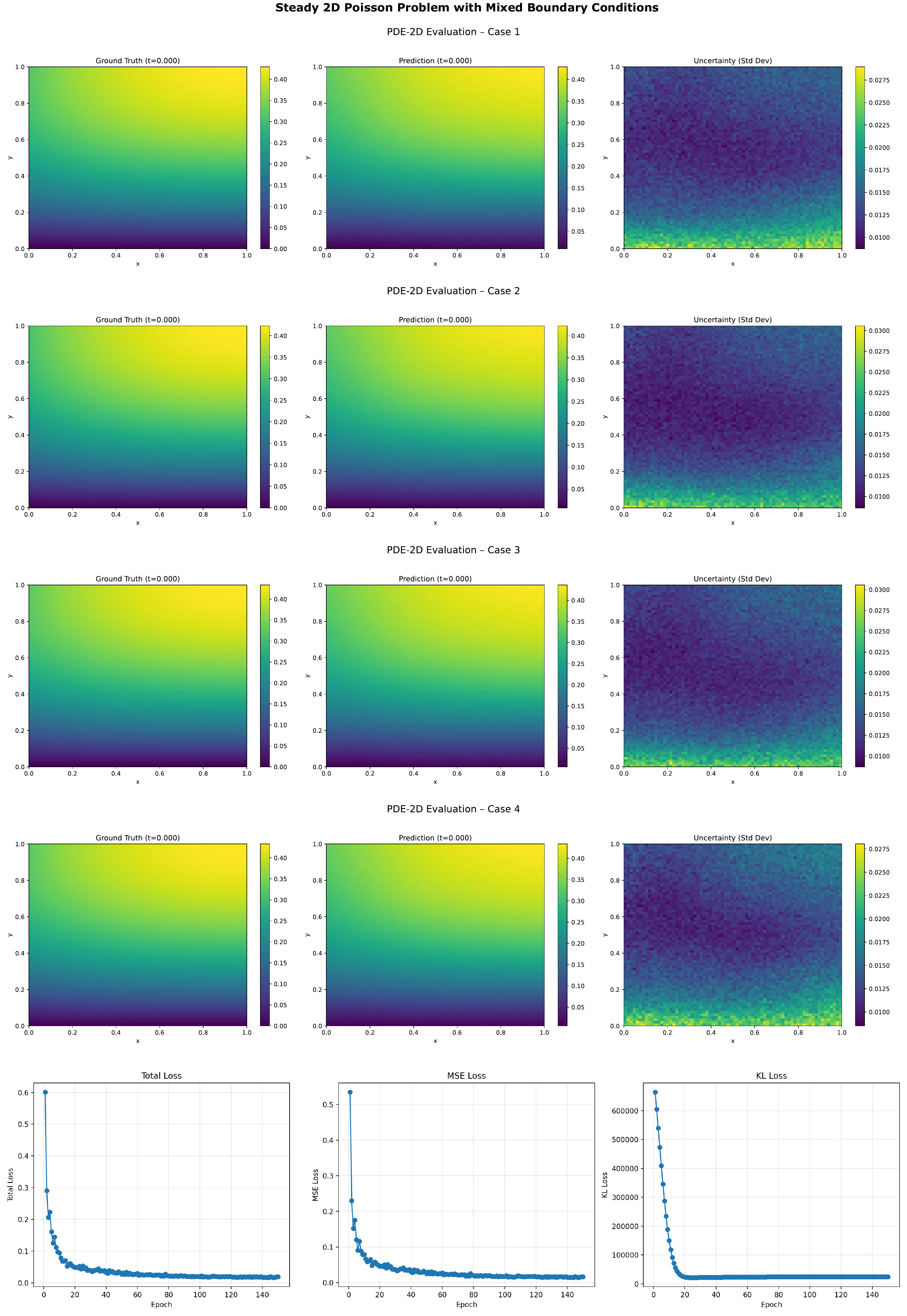}
    \caption{Steady 2D Poisson evaluation with Bayesian DeepONet predictions, uncertainty maps, and training loss histories.}
    \label{fig:steady_2d_poisson}
\end{figure*}

For the steady 2D Poisson problem, the solution is shaped by elliptic diffusion, the source term, and mixed boundary conditions. Figure~\ref{fig:steady_2d_poisson} shows that the Bayesian DeepONet reproduces the main spatial structure of the solver reference across the evaluated cases. The uncertainty stays low over most of the domain, with higher spread near boundary-influenced regions and stronger gradients.

\FloatBarrier
\clearpage

\subsubsection*{\underline{\textbf{PROBLEM 6: Transient 1D Linear Advection Problem}}}

\begin{figure*}[h]
    \centering
    \includegraphics[width=\textwidth]{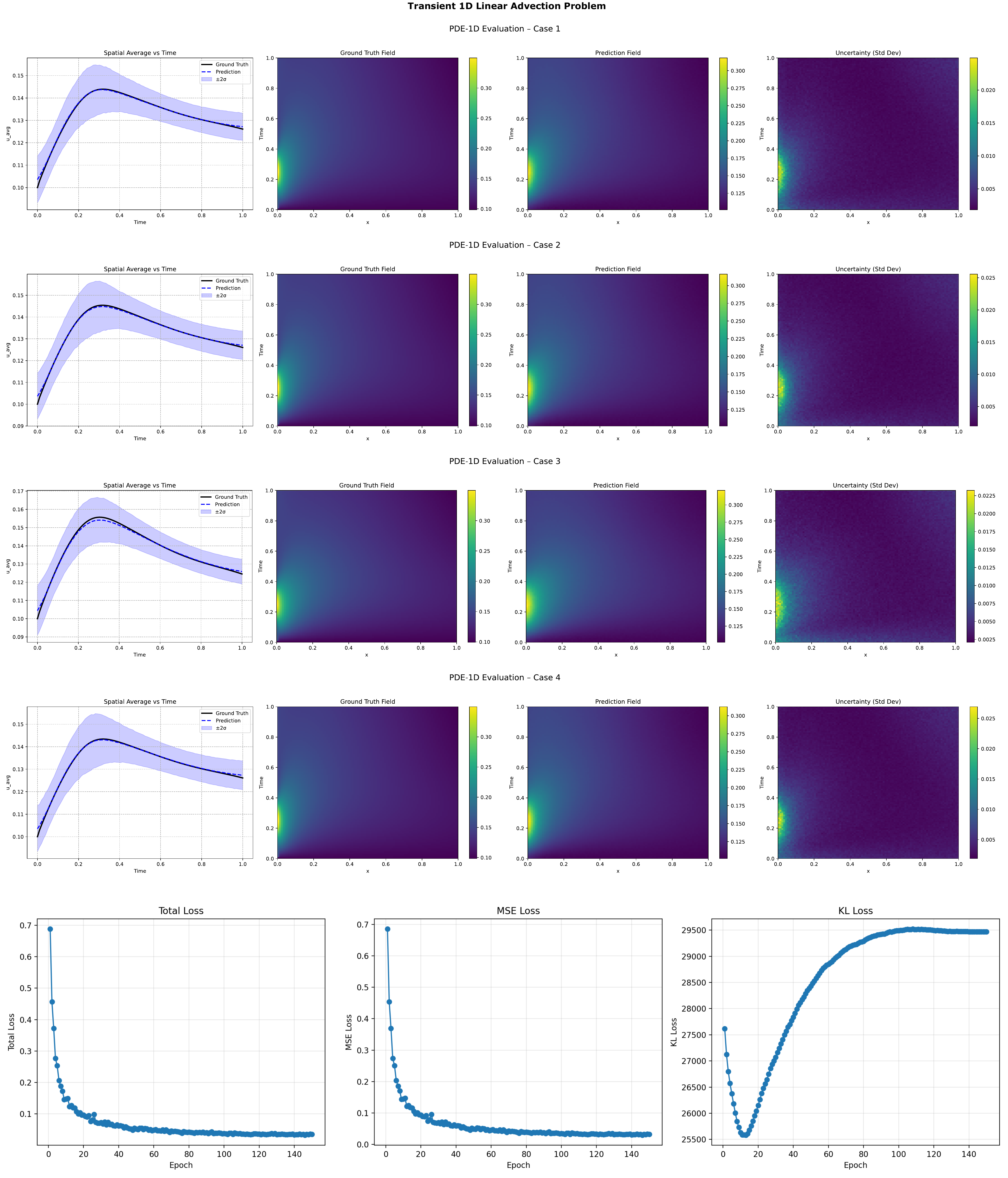}
    \caption{Bayesian DeepONet evaluation for the transient 1D linear advection problem.}
    \label{fig:transient_1d_linear_advection}
\end{figure*}

For the transient 1D linear advection problem, the solution is dominated by transport of the initial profile along the spatial domain. Figure~\ref{fig:transient_1d_linear_advection} shows that the Bayesian DeepONet captures the main movement of the field over the $(x,t)$ grid and follows the spatial-average trend across the evaluated cases. The uncertainty remains small over most regions, with larger spread near sharper transported features. This indicates that the model learns the dominant advection behaviour, while reflecting added uncertainty where the profile changes more abruptly.

\FloatBarrier
\clearpage

\subsubsection*{\underline{\textbf{PROBLEM 7: Transient 2D Gaussian-Jet Heat Diffusion}}}

\begin{figure*}[h]
    \centering
    \includegraphics[width=\textwidth]{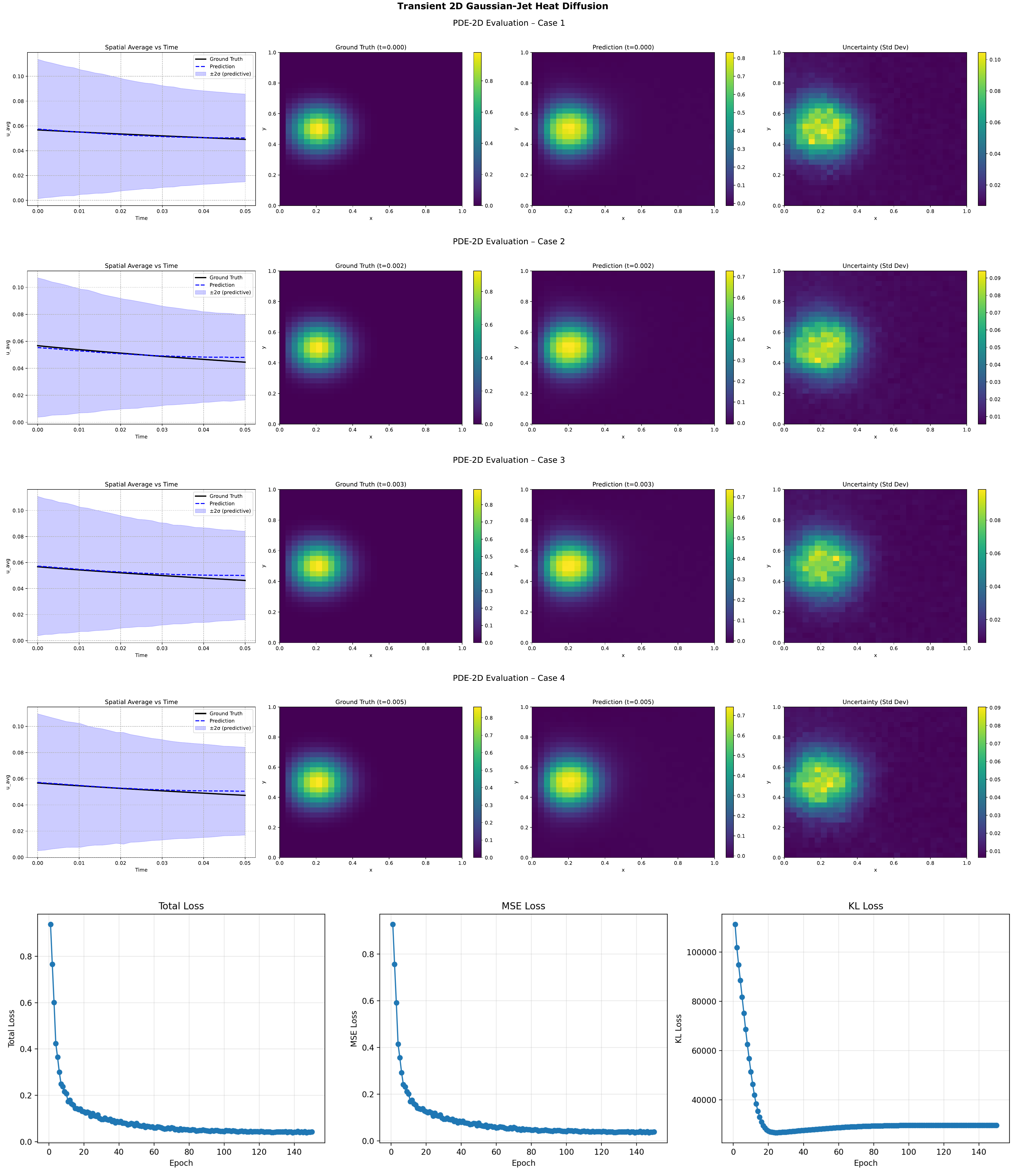}
    \caption{Transient 2D Gaussian-jet heat diffusion evaluation with Bayesian DeepONet predictions, uncertainty maps, and training loss histories.}
    \label{fig:transient_2d_heat_gaussian_jet}
\end{figure*}

For the transient 2D Gaussian-jet heat diffusion problem, the physics is governed by heat spreading from a localized initial pulse over the spatial domain. Figure~\ref{fig:transient_2d_heat_gaussian_jet} shows that the Bayesian DeepONet captures the main diffusion pattern and the decay of the heated region over time. The predictive uncertainty is expected to be higher near the Gaussian jet, where the temperature field changes most sharply. This indicates that the model approximates the transient heat dynamics well while reflecting uncertainty in the high-gradient region.

\FloatBarrier

\begin{table}[t]
\centering
\caption{Training summary for benchmark ODE/PDE operator-learning tasks. The table reports the problem setup, domain and grid resolution, boundary or initial conditions, best training epoch, best training loss, best-checkpoint MSE, and final-epoch MSE.}
\label{tab:operator_training_results}

\scriptsize
\setlength{\tabcolsep}{2.0pt}
\renewcommand{\arraystretch}{1.35}

\rowcolors{2}{gray!18}{white}
\begin{tabularx}{\linewidth}{@{}
>{\centering\arraybackslash}p{0.38cm}
L
>{\raggedright\arraybackslash}p{1.55cm}
L
>{\centering\arraybackslash}p{0.82cm}
>{\centering\arraybackslash}p{0.82cm}
>{\centering\arraybackslash}p{0.75cm}
>{\centering\arraybackslash}p{0.75cm}
@{}}
\toprule
\textbf{No.} &
\textbf{\makecell[l]{Benchmark\\Equation}} &
\textbf{\makecell[l]{Domain /\\Grid}} &
\textbf{\makecell[l]{BC / IC\\Specification}} &
\textbf{\makecell{Best\\Train\\Epoch}} &
\textbf{\makecell{Best\\Train\\Loss}} &
\textbf{\makecell{Best\\MSE}} &
\textbf{\makecell{Final\\MSE}} \\
\midrule

1 &
\textbf{"Steady 1D Parametric Helmholtz Problem"}\newline
$u_{xx}+k_{\mathrm{sq}}u+f(x)=0$\newline
where,\newline
$f(x)=\sin(\pi x)e^{-0.5x^2}$\newline
$k_{\mathrm{sq}}\sim\mathrm{LogNormal}(0.7,0.35)$
\newline
&
$x\in[-3,3]$\newline
\newline
$N_x=240$
&
\textbf{BC:}\newline
\textit{left}: Dirichlet, $u(-3)=0$\newline
\textit{right}: Dirichlet, $u(3)=0$\newline
&
136
&
0.031
&
0.0282
&
0.0293 \\

2 &
\textbf{``Transient 1D Reaction--Diffusion Heat Transfer with Robin Boundaries''}\newline
$u_t=\alpha u_{xx}+\lambda u+f(x)$\newline
where,\newline
$\alpha=0.04$\newline
$f(x)=0.15e^{-50(x-0.4)^2}$\newline
$\lambda\sim\mathrm{Uniform}(-0.8,-0.2)$
\newline
&
$x\in[0,1]$,\newline
$t\in[0,2]$,\newline
\newline
$N_x=160$,\newline
$N_t=200$
&
\textbf{BC:}\newline
\textit{left}: Robin,\newline
$u(0,t)+0.15u_x(0,t)=0.2$\newline
\newline
\textit{right}: Robin,\newline
$u(1,t)+0.25u_x(1,t)=0.05$\newline
\newline
\textbf{IC:}\newline
$u(x,0)=0.2+0.3\sin(\pi x)$\newline
&
150
&
0.0299
&
0.027 
&
0.027 \\

3 &
\textbf{``Parametric Scalar Exponential Growth ODE''}\newline
$u_t=\gamma u$\newline
where,\newline
$\gamma\sim\mathrm{Uniform}(0.2,1.5)$
\newline
&
$t\in[0,6]$\newline
\newline
$N_t=180$
&
\textbf{IC:}\newline
$u(0)=1$
&
145
&
0.0564
&
0.0546
&
0.0796 \\

4 &
\textbf{"Steady 2D Anisotropic Cross-Diffusion Problem"}\newline
$a_xu_{xx}+a_yu_{yy}+\kappa_{xy}u_{xy}+\kappa_{yx}u_{yx}+f(x,y)=0$\newline
where,\newline
$a_x=0.2,\quad a_y=0.1$\newline
$f(x,y)=2\pi^2\sin(\pi x)\sin(\pi y)$\newline
$\kappa_{xy},\kappa_{yx}\sim\mathrm{TruncNormal}(0,0.15;[-0.2,0.2])$
\newline
&
$x\in[0,1]$,\newline
$y\in[0,1]$,\newline
\newline
$N_x=72$,\newline
$N_y=72$
&
\textbf{BC:}\newline
\textit{left}: Dirichlet, $u(0,y)=0$\newline
\textit{right}: Dirichlet, $u(1,y)=0$\newline
\textit{bottom}: Dirichlet, $u(x,0)=0$\newline
\textit{top}: Dirichlet, $u(x,1)=0$
&
139
&
0.1095
&
0.1052
&
0.1132 \\

5 &
\textbf{``Steady 2D Poisson Problem with Mixed Boundary Conditions''}\newline
$k u_{xx}+k u_{yy}+f=0$\newline
where,\newline
$k=1.0,\quad f=1.0$\newline
\newline
&
$x\in[0,1]$,\newline
$y\in[0,1]$,\newline
\newline
$N_x=72$,\newline
$N_y=72$
&
\textbf{BC:}\newline
\textit{left}: Robin,\newline
$u(0,y)+b(y)u_x(0,y)=0$\newline
\newline
where, \newline
$b(y)=1+A\sin(\pi y)$, \newline
$A\sim\mathrm{Uniform}(0.2,0.8)$\newline
\newline
\textit{right}: Neumann, $u_x(1,y)=0$\newline
\textit{bottom}: Dirichlet, $u(x,0)=0$\newline
\textit{top}: Neumann, $u_y(x,1)=0$
&
144
&
0.017
&
0.0146
&
0.0166 \\

6 &
\textbf{``Steady 1D Variable-Coefficient Advection--Diffusion--Reaction Problem''}\newline
$\kappa(x)u_{xx}+cu_x+\lambda u+f(x)=0$\newline
where,\newline
$\kappa(x)=-0.05\left(1+A_{\kappa}\sin(\pi x)\right)$\newline
$A_{\kappa}\sim\mathrm{Uniform}(0.1,0.4)$\newline
$c=-1.0$\newline
$f(x)=x(1-x)$\newline
$\lambda\sim\mathrm{Uniform}(-1.0,0.0)$
\newline
&
$x\in[0,1]$\newline
\newline
$N_x=160$
&
\textbf{BC:}\newline
\textit{left}: Dirichlet, $u(0)=1$\newline
\textit{right}: Neumann, $u_x(1)=0.5$
&
142
&
0.0212
&
0.0185
&
0.0192 \\

7 &
\textbf{``Transient 2D Transport-Dominated Advection--Diffusion Problem''}\newline
$u_t=d u_{xx}+d u_{yy}+c_xu_x+c_yu_y$\newline
where,\newline
$d=0.01$\newline
$c_x\sim\mathrm{LogNormal}(0.0,0.4)$\newline
$c_y\sim\mathrm{LogNormal}(-0.3,0.2)$
\newline
&
$x\in[0,2]$,\newline
$y\in[0,1]$,\newline
$t\in[0,1.5]$\newline
\newline
$N_x=30$,\newline
$N_y=30$,\newline
$N_t=20$
&
\textbf{BC:}\newline
\textit{left}: Dirichlet, $u(0,y,t)=0$\newline
\textit{right}: Neumann, $u_x(2,y,t)=0$\newline
\textit{bottom}: Neumann, $u_y(x,0,t)=0$\newline
\textit{top}: Dirichlet, $u(x,1,t)=\sin(\pi x)$\newline
\newline
\textbf{IC:}\newline
$u(x,y,0)=e^{-20((x-1)^2+(y-0.5)^2)}$
&
108
&
0.1977
&
0.1947
&
0.2083 \\
\bottomrule
\end{tabularx}
\end{table}

\clearpage

\begin{table}[t]
\centering
\caption{Training summary for benchmark ODE/PDE operator-learning tasks, continued.}
\label{tab:operator_training_results_continued}

\scriptsize
\setlength{\tabcolsep}{2.0pt}
\renewcommand{\arraystretch}{1.35}

\rowcolors{2}{gray!18}{white}
\begin{tabularx}{\linewidth}{@{}
>{\centering\arraybackslash}p{0.38cm}
L
>{\raggedright\arraybackslash}p{1.55cm}
L
>{\centering\arraybackslash}p{0.82cm}
>{\centering\arraybackslash}p{0.82cm}
>{\centering\arraybackslash}p{0.75cm}
>{\centering\arraybackslash}p{0.75cm}
@{}}
\toprule
\textbf{No.} &
\textbf{\makecell[l]{Benchmark\\Equation}} &
\textbf{\makecell[l]{Domain /\\Grid}} &
\textbf{\makecell[l]{BC / IC\\Specification}} &
\textbf{\makecell{Best\\Train\\Epoch}} &
\textbf{\makecell{Best\\Train\\Loss}} &
\textbf{\makecell{Best\\MSE}} &
\textbf{\makecell{Final\\MSE}} \\
\midrule

8 &
\textbf{``Transient 1D Linearized Fisher--KPP Reaction--Diffusion''}\newline
$u_t=D(x)u_{xx}+ru+f(x)$\newline
where,\newline
$D(x)=D_0\left(1+A_D\sin(\pi x/3)\right)$\newline
$D_0\sim\mathrm{LogNormal}(-2.6,0.25)$\newline
$A_D\sim\mathrm{Beta}(2.0,8.0)$\newline
$r\sim\mathrm{Normal}(0.5,0.15)$\newline
$f(x)=0.02\sin(\pi x/3)e^{-0.25x^2}$
\newline
&
$x\in[-3,3]$,\newline
$t\in[0,1]$\newline
\newline
$N_x=240$,\newline
$N_t=120$
&
\textbf{BC:}\newline
\textit{left}: Dirichlet, $u(-3,t)=B_L$,\newline
$B_L\sim\mathrm{Beta}(2.0,6.0)$\newline
\newline
\textit{right}: Neumann, $u_x(3,t)=G_R$,\newline
$G_R\sim\mathrm{LogNormal}(-2.3,0.3)$\newline
\newline
\textbf{IC:}\newline
$u(x,0)=0.05+A_0e^{-\frac{(x-x_0)^2}{2w_0^2}}+A_1\sin(\pi x/3)$\newline
$A_0\sim\mathrm{Beta}(2.0,5.0)$,\newline
$x_0\sim\mathrm{Normal}(0,0.6)$,\newline
$w_0\sim\mathrm{LogNormal}(-1.2,0.25)$,\newline
$A_1\sim\mathrm{Normal}(0,0.03)$
\newline
&
141
&
0.5443
&
0.5404
&
0.5407 \\

9 &
\textbf{"Forced Temporal Reaction ODE"}\newline
$u_t=\gamma u+F(t)$\newline
where,\newline
$F(t)=0.5t$\newline
$\gamma\sim\mathrm{Uniform}(-5.0,5.0)$
\newline
&
$t\in[0,1]$\newline
\newline
$N_t=100$
&
\textbf{IC:}\newline
$u(0)=0$
&
140
&
0.0255
&
0.023
&
0.0258 \\

10 &
\textbf{``Transient 1D Linear Advection Problem''}\newline
$u_t=c\,u_x+f(x)$\newline
where,\newline
$f(x)=0.05\sin(2\pi x)$\newline
$c\sim\mathrm{Normal}(-0.8,0.15)$
\newline
&
$x\in[0,1]$,\newline
$t\in[0,1]$\newline
\newline
$N_x=100$,\newline
$N_t=100$
&
\textbf{BC:}\newline
\textit{left}: Dirichlet, $u(0,t)=0.1$\newline
\textit{right}: Neumann, $u_x(1,t)=-0.01$\newline
\newline
\textbf{IC:}\newline
$u(x,0)=0.15+0.2e^{-60(x-0.25)^2}$
\newline
&
146
&
0.0328
&
0.0299
&
0.0323 \\

11 &
\textbf{``Transient 2D Gaussian-Jet Heat Diffusion''}\newline
$u_t=D u_{xx}+D u_{yy}$\newline
where,\newline
$D\sim\mathrm{Uniform}(0.1,0.2)$
\newline
&
$x\in[0,1]$,\newline
$y\in[0,1]$,\newline
$t\in[0,0.05]$\newline
\newline
$N_x=30$,\newline
$N_y=30$,\newline
$N_t=30$
&
\textbf{BC:}\newline
\textit{left}: Dirichlet, $u(0,y,t)=0$\newline
\textit{right}: Dirichlet, $u(1,y,t)=0$\newline
\textit{bottom}: Dirichlet, $u(x,0,t)=0$\newline
\textit{top}: Dirichlet, $u(x,1,t)=0$\newline
\newline
\textbf{IC:}\newline
$u(x,y,0)=e^{-\frac{(x-0.2)^2+(y-0.5)^2}{2(0.1)^2}}$
&
139
&
0.0389
&
0.0359
&
0.0396 \\

\bottomrule
\end{tabularx}
\end{table}

\subsection{Runtime Analysis}
\label{sec:runtime_analysis}

This section reports the per-sample runtime cost of solver-backed data generation and compares it with neural-operator inference across the benchmark problems. All runtimes were measured in a 64-bit x86 virtualised environment with 8 online CPUs on an AMD EPYC 7763 processor, using 4 CPU cores with 2 threads per core, a single NUMA node, and a 32 MiB L3 cache. Table~\ref{tab:runtime_inference_speedup} reports the data-generation time per sample, total prediction time, pure model-forward time, and the corresponding speedup factors. System speedup is computed as data-generation time per sample divided by total prediction time per sample, while inference speedup is computed as data-generation time per sample divided by pure model-forward time per sample. This separates end-to-end inference overhead from the neural operator evaluation itself.

\begin{table*}[htbp]
\centering
\caption{Runtime summary for per-sample data generation, neural-operator inference, and speedup.}
\label{tab:runtime_inference_speedup}
\scriptsize
\renewcommand{\arraystretch}{1.22}
\setlength{\tabcolsep}{4pt}

\begin{tabularx}{\textwidth}{@{}
    >{\raggedright\arraybackslash}X
    >{\centering\arraybackslash}p{1.75cm}
    >{\centering\arraybackslash}p{1.65cm}
    >{\centering\arraybackslash}p{1.75cm}
    >{\centering\arraybackslash}p{1.45cm}
    >{\centering\arraybackslash}p{1.55cm}
@{}}
\toprule
\textbf{Problem} &
\textbf{\makecell{Data Gen.\\per Sample (s)}} &
\textbf{\makecell{Total Pred.\\Time (s)}} &
\textbf{\makecell{Model Forward\\Time (s)}} &
\textbf{\makecell{System\\Speedup}} &
\textbf{\makecell{Inference\\Speedup}} \\
\midrule

Transient 1D Heat Transfer &
1.769 & 0.0635 & 0.0555 & 27.87 & 31.85 \\

Transient 1D Linearized Fisher--KPP &
0.452 & 0.0621 & 0.0482 & 7.28 & 9.38 \\

Parametric Scalar Exponential Growth ODE &
0.0008 & 0.0010 & 0.0006 & 0.73 & 1.24 \\

Forced Temporal Reaction ODE &
0.0010 & 0.0023 & 0.0008 & 0.42 & 1.22 \\

Steady 2D Anisotropic Cross-Diffusion &
0.102 & 0.1038 & 0.0992 & 0.98 & 1.03 \\

Steady 2D Poisson &
0.112 & 0.1290 & 0.1254 & 0.87 & 0.90 \\

Steady 1D Parametric Helmholtz &
0.0229 & 0.0046 & 0.0011 & 4.98 & 20.17 \\

Steady 1D Advection--Diffusion--Reaction &
0.0267 & 0.0074 & 0.0010 & 3.61 & 27.07 \\

Transient 1D Linear Advection &
0.244 & 0.0172 & 0.0101 & 14.20 & 24.16 \\

Transient 2D Transport &
0.128 & 0.0845 & 0.0771 & 1.51 & 1.66 \\

Transient 2D Gaussian-Jet Heat Diffusion &
0.171 & 0.1208 & 0.1139 & 1.42 & 1.50 \\

\bottomrule
\end{tabularx}
\end{table*}


\subsection{FEniCSx Finite-Element Solver Backend}
\label{sec:fenicsx_solver_backend}

The FEniCSx solver backend converts the PDE configuration produced by the input and data-generation modules into finite-element solutions on the requested output grid. As shown in Figure~\ref{fig:solver_system_architecture}, the solver receives the PDE operator, coefficient fields, boundary data, initial condition and time configuration. It then detects whether the problem is purely temporal, one-dimensional or two-dimensional. Pure temporal ODEs are sent to a dedicated time integrator, while one-dimensional PDEs are embedded as thin two-dimensional strips with homogeneous Neumann conditions on inactive boundaries.

\subsubsection{Finite-Element Weak Form Assembly}
\label{sec:fem_weak_form_assembly}

For spatial ODE/PDE problems, the solver starts from the general linear transient form
\begin{equation}
    \frac{\partial u}{\partial t}
    =
    \nabla\cdot\left(K(\mathbf{x})\nabla u\right)
    +
    \boldsymbol{\alpha}(\mathbf{x})\cdot\nabla u
    +
    \gamma(\mathbf{x})u
    +
    f(\mathbf{x},t),
    \qquad
    \mathbf{x}\in\Omega,\; t\in[0,T].
    \label{eq:solver_strong_form}
\end{equation}
For steady problems, the time-derivative term is omitted. In two dimensions, the diffusion tensor and advection field are written as
\begin{equation}
    K(\mathbf{x})
    =
    \begin{bmatrix}
        \beta_x(\mathbf{x}) & \eta_{xy}(\mathbf{x}) \\
        \eta_{yx}(\mathbf{x}) & \beta_y(\mathbf{x})
    \end{bmatrix},
    \qquad
    \boldsymbol{\alpha}(\mathbf{x})
    =
    \begin{bmatrix}
        \alpha_x(\mathbf{x}) \\
        \alpha_y(\mathbf{x})
    \end{bmatrix}.
\end{equation}

The solver builds a rectangular FEniCSx mesh $\mathcal{T}_h$ over the requested domain and uses continuous piecewise-linear finite elements,
\begin{equation}
    V_h = \mathrm{CG}_1(\mathcal{T}_h).
\end{equation}
The finite-element solution is approximated as
\begin{equation}
    u_h(\mathbf{x})
    =
    \sum_{j=1}^{N_h}
    U_j\phi_j(\mathbf{x}),
\end{equation}
where $\{\phi_j\}_{j=1}^{N_h}$ are basis functions and $U_j$ are the unknown degrees of freedom. Scalar, callable, and array-valued coefficients from the input specification are converted into FEniCSx constants or finite-element functions before assembly.

To obtain the weak form, Eq.~\eqref{eq:solver_strong_form} is multiplied by a test function $v\in V_h$ and the diffusion term is integrated by parts. The spatial bilinear form assembled by the solver is
\begin{equation}
    a(u,v)
    =
    \int_{\Omega}
    (K\nabla u)\cdot\nabla v\,d\mathbf{x}
    -
    \int_{\Omega}
    \left(\boldsymbol{\alpha}\cdot\nabla u\right)v\,d\mathbf{x}
    -
    \int_{\Omega}
    \gamma uv\,d\mathbf{x}
    +
    a_{\partial\Omega}(u,v),
    \label{eq:solver_bilinear_form}
\end{equation}
and the linear form is
\begin{equation}
    L(v)
    =
    \int_{\Omega}
    f v\,d\mathbf{x}
    +
    L_{\partial\Omega}(v).
    \label{eq:solver_linear_form}
\end{equation}
Here, the first term in $a(u,v)$ is the diffusion contribution, the second is the advection contribution, and the third is the reaction contribution. The terms $a_{\partial\Omega}$ and $L_{\partial\Omega}$ collect boundary contributions.

Dirichlet boundary conditions are imposed directly on boundary degrees of freedom. Neumann and Robin conditions are added through boundary integrals on tagged edges. With the solver convention
\begin{equation}
    q = -\mathbf{n}\cdot K\nabla u,
\end{equation}
a Neumann edge $\Gamma_e$ with prescribed flux $q=g_e$ contributes
\begin{equation}
    L_{\Gamma_e}^{N}(v)
    =
    \int_{\Gamma_e}
    g_e v\,ds .
\end{equation}
For a Robin condition on $\Gamma_e$,
\begin{equation}
    a_e u + b_e q = c_e,
\end{equation}
the corresponding bilinear and linear boundary terms are
\begin{equation}
    a_{\Gamma_e}^{R}(u,v)
    =
    \int_{\Gamma_e}
    \frac{a_e}{b_e}uv\,ds,
    \qquad
    L_{\Gamma_e}^{R}(v)
    =
    \int_{\Gamma_e}
    \frac{c_e}{b_e}v\,ds .
\end{equation}
These terms are added only on edges where Robin conditions are specified.

After substituting the finite-element expansion into the weak form and testing with each basis function $\phi_i$, the solver obtains
\begin{equation}
    \sum_{j=1}^{N_h}
    A_{ij}U_j
    =
    b_i,
    \qquad
    A_{ij}=a(\phi_j,\phi_i),
    \qquad
    b_i=L(\phi_i).
\end{equation}
In matrix form,
\begin{equation}
    A U = b .
\end{equation}
Thus, the bilinear form defines the matrix $A$, the linear form defines the right-hand side $b$, and the vector $U$ stores the finite-element coefficients of the solution.

For transient problems, the spatial weak form is combined with the mass form
\begin{equation}
    m(u,v)
    =
    \int_{\Omega}
    uv\,d\mathbf{x}.
\end{equation}
For example, a backward-Euler step from $u_h^n$ to $u_h^{n+1}$ is assembled as
\begin{equation}
    \frac{1}{\Delta t}
    m(u_h^{n+1},v)
    +
    a(u_h^{n+1},v)
    =
    \frac{1}{\Delta t}
    m(u_h^n,v)
    +
    L(v).
\end{equation}
Other time-integration schemes change the time-discretisation coefficients, but reuse the same spatial weak-form components. After solving the finite-element system, the solution is evaluated on the requested uniform output grid and stored as $U(y,x)$ for steady problems or $U(t,y,x)$ for transient problems.

\subsubsection{FEniCSx Solver Workflow}
\label{sec:fenicsx_solver_workflow}

\begin{figure*}[htbp]
    \centering
    \includegraphics[width=\textwidth]{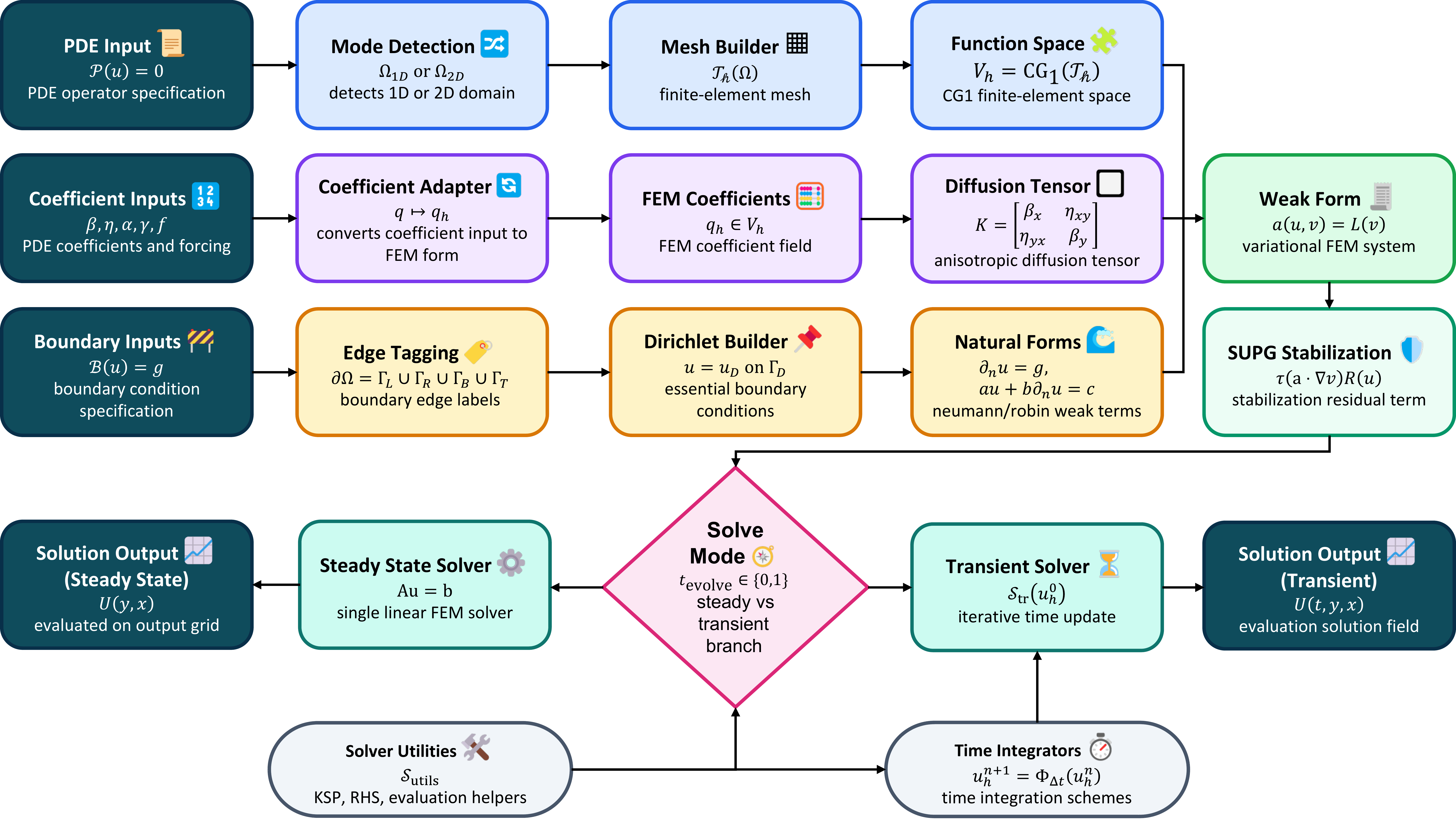}
    \caption{FEniCSx solver workflow for converting parsed ODE/PDE inputs into finite-element solutions. The solver detects the problem mode, builds the mesh and function space, converts coefficients and boundary data into FEM forms, assembles the weak problem with optional SUPG stabilisation, and returns either a steady solution field or a transient trajectory on the requested output grid.}
    \label{fig:solver_system_architecture}
\end{figure*}

The solver workflow converts a sampled problem configuration into a finite-element solution and then evaluates that solution on the requested output grid. As shown in Figure~\ref{fig:solver_system_architecture}, the solver receives the parsed PDE operator, sampled coefficient and forcing terms, boundary-condition specification, and, when required, the initial condition and time configuration. The mode detector first decides whether the problem is a pure temporal ODE, a one-dimensional spatial problem, or a two-dimensional PDE. Pure temporal ODEs are sent directly to the time integrator. One-dimensional PDEs are embedded as thin two-dimensional strips, with inactive boundaries set to homogeneous Neumann conditions, so that the same FEniCSx assembly path can be reused. The problem classes routed through this workflow are summarised in Table~\ref{tab:dataset_supported_problem_classes}.

For spatial problems, the mesh builder creates a rectangular finite-element mesh over the configured domain. The solver then defines a first-order continuous Galerkin space and converts scalar, callable, and array-valued coefficients into FEniCSx-compatible constants or functions. The coefficient adapter builds the diffusion tensor from the directional and cross-diffusion terms. Boundary edges are tagged as bottom, top, left, and right, allowing Dirichlet, Neumann, and Robin conditions to be assigned independently on each edge.

The boundary-condition stage separates essential and natural boundary terms. Dirichlet values are imposed directly on boundary degrees of freedom. Neumann and Robin conditions are added to the weak form through edge-specific boundary integrals. The solver may also add a SUPG stabilisation term before assembly. This gives a single variational system containing the PDE operator, coefficient fields, source term, boundary contributions, and optional stabilisation.

The final solve path depends on whether the problem is steady or transient. For steady problems, the solver assembles one sparse linear system and applies a direct linear solve. For transient problems, the same spatial weak-form components are passed to the selected time integrator; the supported schemes are listed in Table~\ref{tab:data_generation_supported_features}. Time-dependent source terms are updated at each step when needed. After the solve, the finite-element solution is evaluated on the uniform output grid and returned as a steady field $U(y,x)$ or a transient field $U(t,y,x)$.

\begin{table}[htbp]
\centering
\caption{Problem classes handled by the data-generation and solver workflow.}
\label{tab:dataset_supported_problem_classes}
\small
\renewcommand{\arraystretch}{1.2}
\setlength{\tabcolsep}{5pt}
\begin{tabularx}{0.95\linewidth}{
    >{\raggedright\arraybackslash}p{0.17\linewidth}
    >{\raggedright\arraybackslash}X
    >{\raggedright\arraybackslash}p{0.08\linewidth}
}
\toprule
\rowcolor{gray!12}
\textbf{Problem Class} & \textbf{General Form} & \textbf{Query Coordinate} \\
\midrule

Temporal ODE &
$
u_t
=
\gamma(t)u + f(t)
$
&
$t$ \\

Spatial ODE &
$
\beta_x u_{xx}
+
\alpha_x u_x
+
\gamma u
+
f_{\mathrm{const}}
=
0
$
&
$x$ \\

Transient 1D PDE &
$
u_t
=
\beta_x u_{xx}
+
\alpha_x u_x
+
\gamma u
+
f_{\mathrm{const}}
$
&
$(x,t)$ \\

Steady 2D PDE &
$
\beta_x u_{xx}
+
\beta_y u_{yy}
+
\eta_{xy}u_{xy}
+
\eta_{yx}u_{yx}
+
\alpha_x u_x
+
\alpha_y u_y
+
\gamma u
+
f_{\mathrm{const}}
=
0
$
&
$(x,y)$ \\

Transient 2D PDE &
$
u_t
=
\beta_x u_{xx}
+
\beta_y u_{yy}
+
\eta_{xy}u_{xy}
+
\eta_{yx}u_{yx}
+
\alpha_x u_x
+
\alpha_y u_y
+
\gamma u
+
f_{\mathrm{const}}
$
&
$(x,y,t)$ \\

\bottomrule
\end{tabularx}
\end{table}


\subsubsection{Solver Validation Studies}
\label{sec:solver_validation_studies}

We validate the solver using a mix of benchmark ODE/PDE cases and manufactured-solution tests. In the manufactured-solution cases, an analytical solution is chosen first, and the corresponding source term, boundary conditions, and initial condition are derived from it. This gives a controlled reference for checking the assembled weak form, boundary treatment, and time integration. Tables~\ref{tab:validation_dataset_summary}, ~\ref{tab:validation_dataset_summary_continued} and ~\ref{tab:validation_dataset_summary_continued_2} report the equation class, boundary and initial conditions, grid resolution, time scheme, and error metrics. The validation set covers temporal ODEs, steady Poisson and diffusion problems, transient heat equations, advection--diffusion--reaction systems, tensor diffusion, and mixed boundary conditions. The reported $L_2$ and $L_{\infty}$ errors indicate that the solver matches the analytical references within the expected discretisation accuracy for both steady and transient cases. These tests also check that edge-wise Dirichlet, Neumann, and Robin conditions are applied consistently before the solver is used for dataset generation.

\clearpage

\subsubsection*{\underline{\textbf{CASE 1: Transient 1D Diffusion Problem}}}
\begin{figure*}[h]
    \centering
    \includegraphics[width=\textwidth]{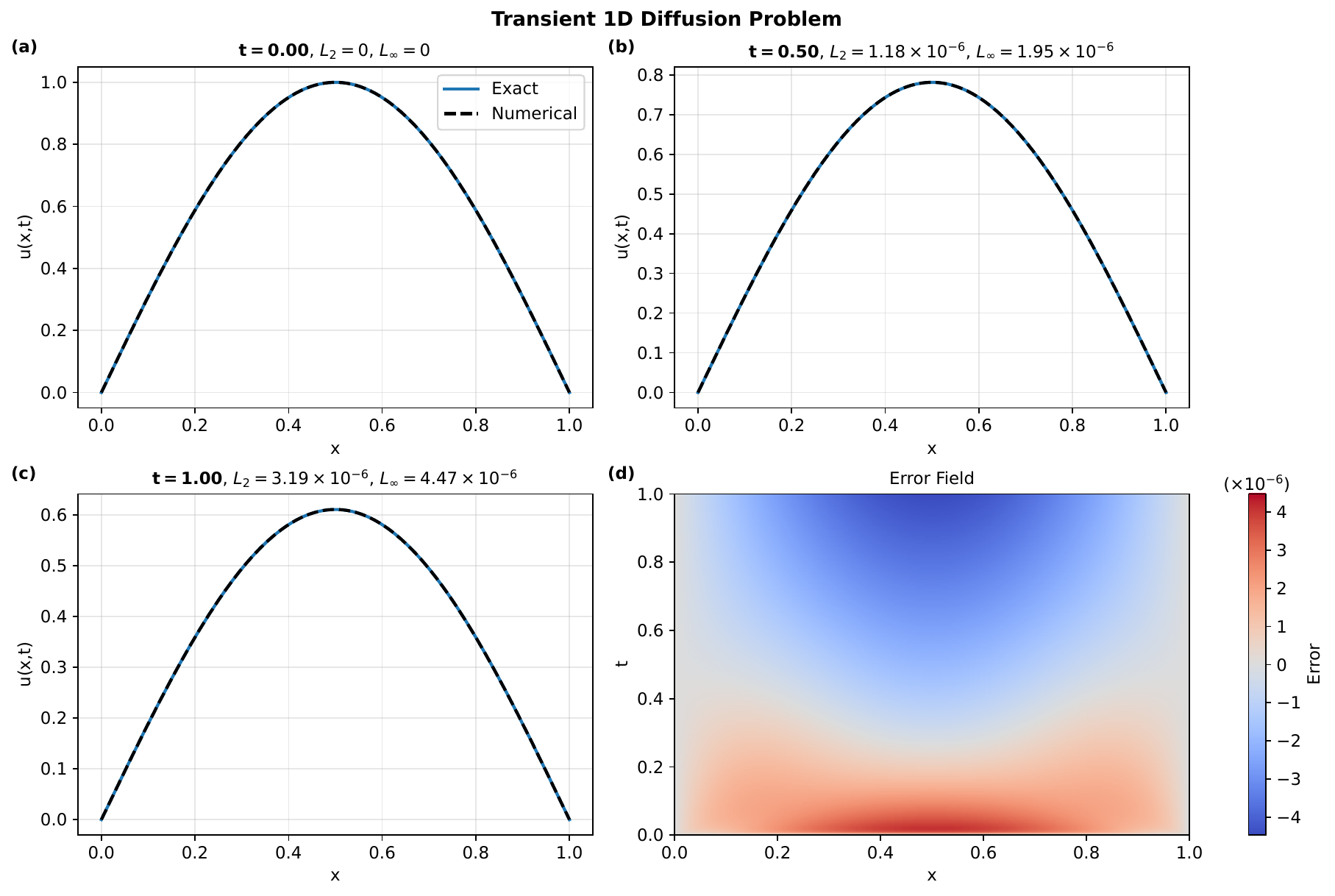}
    \caption{Validation of the transient 1D diffusion solver against the analytical heat-equation solution.}
    \label{fig:validation_transient_1d_pde_diffusion}
\end{figure*}

The transient 1D diffusion solver is validated against the analytical heat-equation solution with homogeneous Dirichlet boundaries. A sine initial condition diffuses over time, showing amplitude decay while boundary values remain fixed. Figure~\ref{fig:validation_transient_1d_pde_diffusion} shows good agreement at selected time instances. The $L_2$ and $L_{\infty}$ errors are on the order of $10^{-6}$, with uniformly small residuals across the $(x,t)$ domain, confirming correct transient evolution and boundary handling.

\FloatBarrier
\subsubsection*{\underline{\textbf{CASE 2: Steady 2D Poisson Problem with Dirichlet Boundary Conditions}}}
\begin{figure}[h]
    \centering
    \includegraphics[width=\textwidth]{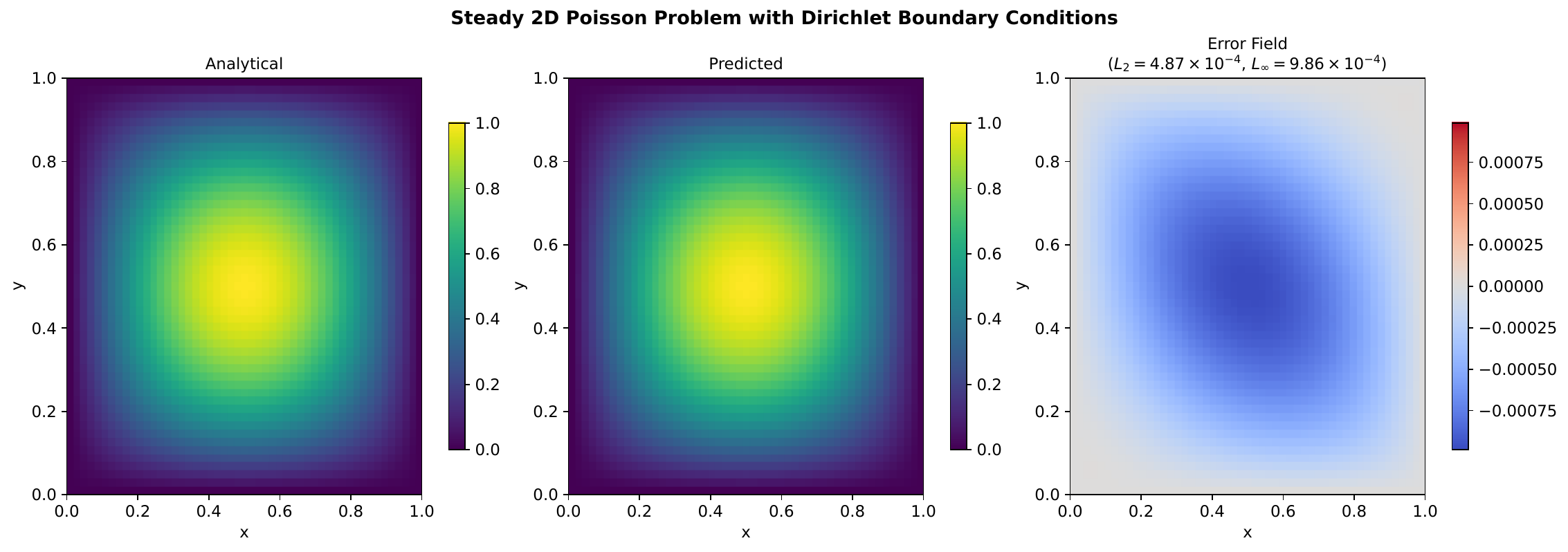}
    \caption{Validation of the steady 2D Poisson solver with homogeneous Dirichlet boundary conditions.}
    \label{fig:validation_steady_poisson_dirichlet_sinpi}
\end{figure}

The steady 2D Poisson solver is tested with homogeneous Dirichlet boundaries and a sinusoidal source-driven solution. The numerical field reproduces the analytical profile, including the central peak and boundary decay, as shown in Figure~\ref{fig:validation_steady_poisson_dirichlet_sinpi}. Errors remain small across the domain, with $L_2=4.87\times10^{-4}$ and $L_{\infty}=9.86\times10^{-4}$, supporting the steady weak-form implementation.

\FloatBarrier
\clearpage
\subsubsection*{\underline{\textbf{CASE 3: Temporal Decay ODE}}}
\begin{figure*}[h]
    \centering
    \includegraphics[width=\textwidth]{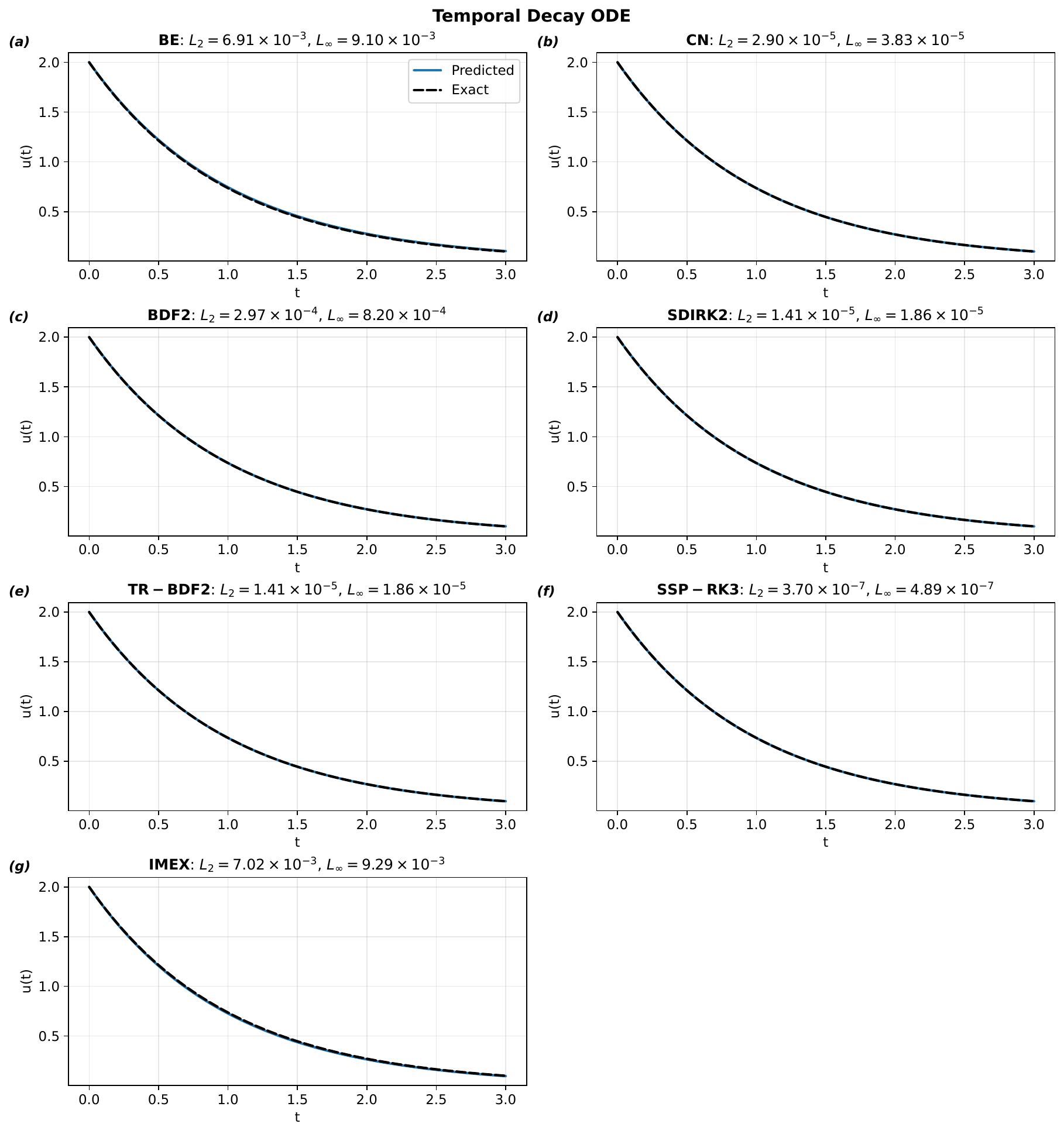}
    \caption{Validation of the temporal decay ODE against the analytical exponential solution.}
    \label{fig:validation_temporal_ode}
\end{figure*}

For the temporal decay ODE, the solver is tested on a first-order system with negative reaction coefficient, where the governing physics is exponential decay from the prescribed initial value. Figure~\ref{fig:validation_temporal_ode} compares the numerical trajectories with the analytical solution across the time interval. The predicted curves closely follow the exact decay profile for the tested time schemes, with small reported $L_2$ and $L_{\infty}$ errors. This validates the pure temporal solve path and confirms that the time integrators preserve the expected decay behaviour.

\FloatBarrier
\subsubsection*{\underline{\textbf{CASE 4: Transient 2D Advection-Diffusion-Reaction Problem}}}
\begin{figure}[h]
    \centering
    \includegraphics[width=0.95\textwidth]{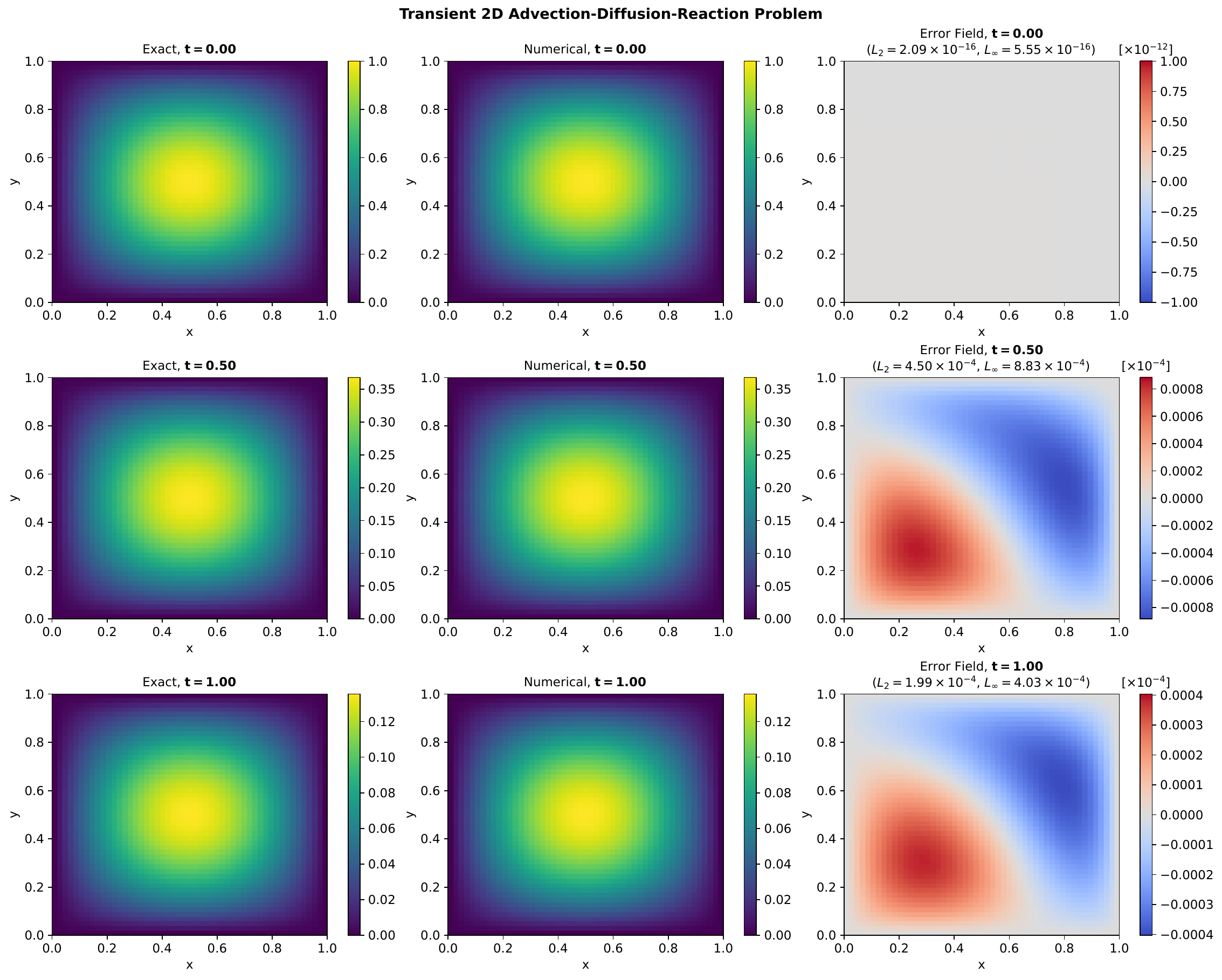}
    \caption{Transient 2D advection--diffusion--reaction validation under homogeneous Dirichlet boundaries.}
    \label{fig:validation_transient_advdiff_react_dirichlet_decay_backward_euler}
\end{figure}

For the transient 2D advection--diffusion--reaction case, the solver is tested on a decaying solution with homogeneous Dirichlet boundary conditions. The governing physics combines diffusion-driven smoothing, advective transport, reaction-driven decay, and forcing. As shown in Figure~\ref{fig:validation_transient_advdiff_react_dirichlet_decay_backward_euler}, the numerical fields closely follow the analytical fields at $t=0$, $t=0.5$, and $t=1.0$. The solution magnitude decreases over time while the boundary values remain fixed at zero. The error fields remain small across the domain, supporting the transient FEM solve path, backward-Euler time stepping, and Dirichlet boundary enforcement.

\FloatBarrier
\subsubsection*{\underline{\textbf{CASE 5: Transient 1D Heat Eigenmode}}}

\begin{figure}[h]
    \centering
    \includegraphics[width=0.75\textwidth]{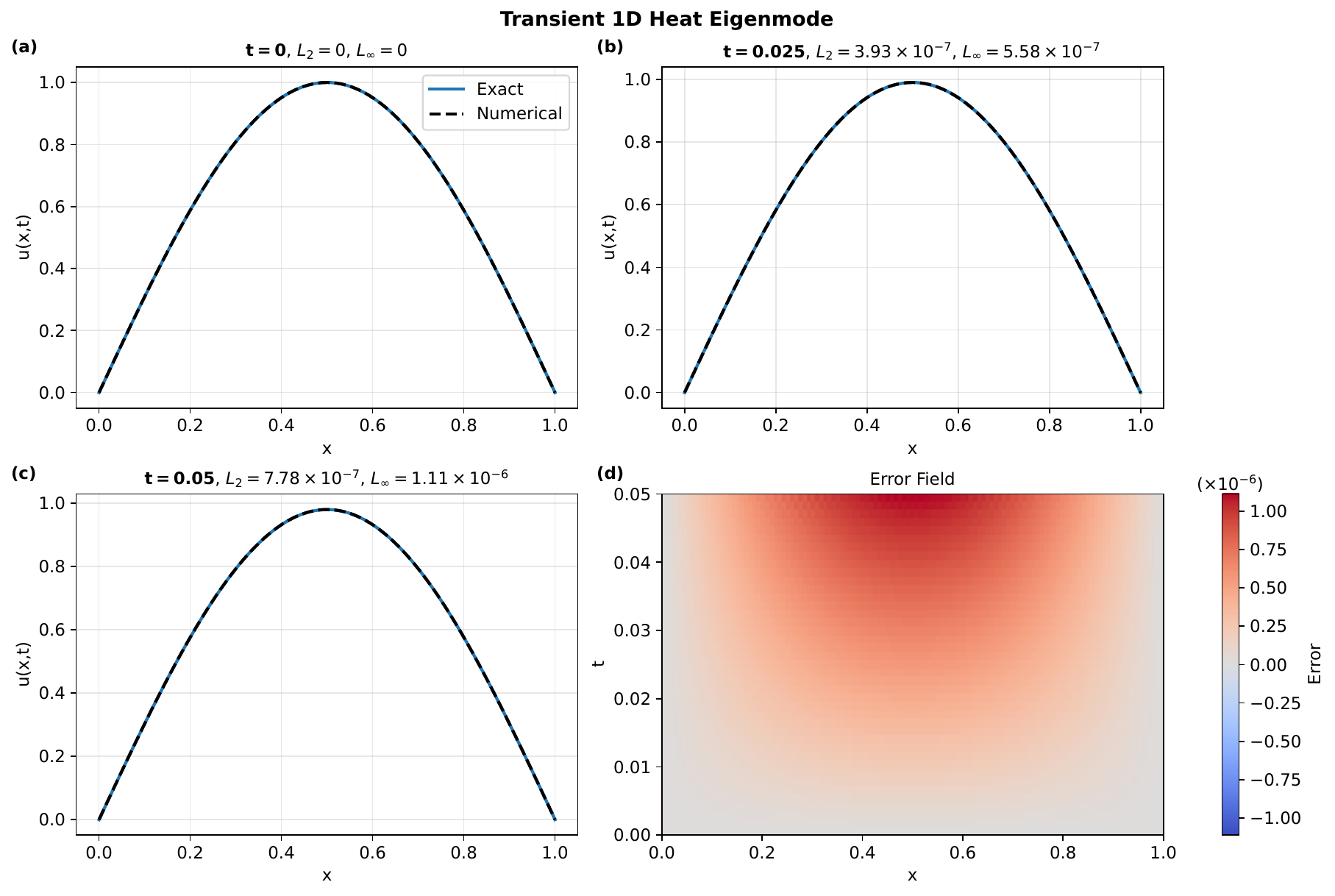}
    \caption{Validation of the transient 1D heat eigenmode against the analytical diffusion solution.}
    \label{fig:transient_1d_heat_eigenmode}
\end{figure}

The solver is tested on the homogeneous heat equation with zero Dirichlet boundaries. A sine initial profile decays in amplitude while preserving its shape. Figure~\ref{fig:transient_1d_heat_eigenmode} shows close agreement with the analytical solution and uniformly small errors across the $(x,t)$ domain.

\FloatBarrier

\begin{table}[htbp]
\centering
\caption{Solver validation results for benchmark ODE/PDE cases (1/3). The table reports equation type, boundary and initial conditions, domain and grid, exact solution, time scheme, and validation errors.}
\label{tab:validation_dataset_summary}

\scriptsize
\setlength{\tabcolsep}{2.0pt}
\renewcommand{\arraystretch}{1.35}

\rowcolors{2}{gray!18}{white}
\begin{tabularx}{\linewidth}{@{}
>{\centering\arraybackslash}p{0.25cm}
>{\raggedright\arraybackslash}p{2.75cm}
>{\raggedright\arraybackslash}p{2.75cm}
>{\raggedright\arraybackslash}p{1.35cm}
L
>{\centering\arraybackslash}p{1.0cm}
>{\centering\arraybackslash}p{1.5cm}
>{\centering\arraybackslash}p{1.5cm}
@{}}
\toprule
\textbf{No.} &
\textbf{\makecell{Equation\\Type}} &
\textbf{\makecell{BC / IC \\Specification}} &
\textbf{\makecell{Domain}} &
\textbf{\makecell{Exact\\Solution}} &
\textbf{\makecell{Time\\Scheme}} &
\textbf{\makecell{$L_2$}} &
\textbf{$L_{\infty}$} \\
\midrule

1 &
\textbf{``Transient 1D Heat Diffusion''}\newline
$u_t=\beta u_{xx}$\newline
where,\newline
$\beta=0.05$
&
\textbf{BC:}\newline
\textit{left}: Dirichlet,\newline $u(0,t)=0$\newline
\textit{right}: Dirichlet,\newline $u(1,t)=0$\newline
\newline
\textbf{IC:}\newline
$u(x,0)=\sin(\pi x)$
&
$x\in[0,1]$\newline
$t\in[0,1]$\newline
\newline
$N_x=201$\newline
$N_t=201$
&
$u(x,t)=e^{-\beta\pi^2t}\sin(\pi x)$
&
TR-BDF2
&
\textbf{t = 0:}\newline
$1.03 \times 10^{-16}$\newline
\textbf{t = 0.5:}\newline
$3.43 \times 10^{-6}$\newline
\textbf{t = 1:}\newline
$4.78 \times 10^{-6}$\newline
&
\textbf{t = 0:}\newline
$2.22 \times 10^{-16}$\newline
\textbf{t = 0.5:}\newline
$5.18 \times 10^{-6}$\newline
\textbf{t = 1:}\newline
$6.72 \times 10^{-6}$\newline \\

2 &
\textbf{``Steady 1D Poisson Diffusion''}\newline
$ \beta u_{xx}+f=0 $\newline
where,\newline
$\beta=1.0$\newline
$f=1.0$
&
\textbf{BC:}\newline
\textit{left}: Dirichlet,\newline $u(0)=0$\newline
\textit{right}: Dirichlet,\newline $u(1)=0$\newline
&
$x\in[0,1]$\newline
\newline
$N_x=401$
&
$u(x)=\frac{1}{2}x(1-x)$
&
--
& 
$3.35 \times 10^{-15}$
& 
$6.02 \times 10^{-15}$ \\

3 &
\textbf{``Temporal Exponential Decay ODE''}\newline
$u_t=\gamma u$\newline
where,\newline
$\gamma=-1.0$
&
\textbf{IC:}\newline
$u(0)=2$
&
$t\in[0,3]$\newline
\newline
$N_t=121$
&
$u(t)=2e^{-t}$
&
\mbox{}\newline
BE \newline
CN \newline
BDF2 \newline
SDIRK2 \newline
TR-BDF2 \newline
SSP-RK3 \newline
IMEX \newline
& 
\textbf{t = 3:}\newline
$6.91 \times 10^{-3}$ \newline
$2.9 \times 10^{-5}$ \newline
$2.97 \times 10^{-4}$ \newline
$1.41 \times 10^{-5}$ \newline
$1.41 \times 10^{-5}$ \newline
$3.7 \times 10^{-7}$ \newline
$7.02 \times 10^{-3}$ \newline

& 
\textbf{t = 3:}\newline
$9.1 \times 10^{-3}$ \newline
$3.83 \times 10^{-5}$ \newline
$8.19 \times 10^{-4}$ \newline
$1.86 \times 10^{-5}$ \newline
$1.86 \times 10^{-5}$ \newline
$4.89 \times 10^{-7}$ \newline
$9.29 \times 10^{-3}$ \newline

\\

4 &
\textbf{``Steady 2D Dirichlet Advection--Diffusion--Reaction''}\newline
$\beta_xu_{xx}+\beta_yu_{yy}-\alpha_xu_x-\alpha_yu_y+\gamma u+f(x,y) = 0$\newline
where,\newline
$\beta_x=\beta_y=0.1$\newline
$\alpha_x=1.0$\newline 
$\alpha_y=0.5$\newline
$\gamma=0.2$
&
\textbf{BC:}\newline
\textit{left}: Dirichlet,\newline 
$u(0,y)=0$\newline
\textit{right}: Dirichlet,\newline
$u(1,y)=0$\newline
\textit{bottom}: Dirichlet,\newline
$u(x,0)=0$\newline
\textit{top}: Dirichlet, \newline
$u(x,1)=0$\newline
&
$x\in[0,1]$\newline
$y\in[0,1]$\newline
\newline
$N_x=51$\newline
$N_y=51$
&
$u(x,y)=\sin(\pi x)\sin(\pi y)$\newline
&
--
&
$1.19 \times 10^{-3}$
& 
$2.57 \times 10^{-3}$ \\

5 &
\textbf{"Transient 2D Neumann  Advection--Diffusion"}\newline
$u_t = \beta_xu_{xx}+\beta_yu_{yy}-\alpha_xu_x-\alpha_yu_y+\gamma u+f$\newline
where,\newline
$\beta_x=1.0$\newline
$\beta_y=0.5$\newline
$\alpha_x=2.0$\newline
$\alpha_y=3.0$\newline
$\gamma=0.0$\newline
$f=6$
&
\textbf{BC:}\newline
\textit{left}: Neumann,\newline
$u_x(0,y,t)=1.0$\newline
\textit{right}: Neumann,\newline
$u_x(1,y,t)=1.0$\newline
\textit{bottom}: Neumann,\newline
$u_y(x,0,t)=1.0$\newline
\textit{top}: Neumann,\newline
$u_y(x,1,t)=1.0$\newline
\newline
\textbf{IC:}\newline
$u(x,y,0)=x+y$
&
$x\in[0,1]$\newline
$y\in[0,1]$\newline
$t\in[0,0.008]$\newline
\newline
$N_x=51$\newline
$N_y=51$\newline
$N_t=101$
&
$u(x,y,t)=x+y+t$
&
\mbox{} \newline
BE \newline
CN \newline
BDF2 \newline
SDIRK2 \newline
TR-BDF2 \newline
SSP-RK3 \newline
IMEX \newline
&
\textbf{t = 0.008:}\newline
$2.17 \times 10^{-4}$ \newline
$2.18 \times 10^{-4}$ \newline
$2.18 \times 10^{-4}$ \newline
$2.18 \times 10^{-4}$ \newline
$2.18 \times 10^{-4}$ \newline
$2.16 \times 10^{-4}$ \newline
$2.17 \times 10^{-4}$ \newline
&
\textbf{t = 0.008:}\newline
$1.06 \times 10^{-3}$ \newline
$1.06 \times 10^{-3}$ \newline
$1.06 \times 10^{-3}$ \newline
$1.06 \times 10^{-3}$ \newline
$1.06 \times 10^{-3}$ \newline
$1.05 \times 10^{-3}$ \newline
$1.06 \times 10^{-3}$ \newline \\

6 &
\textbf{``Transient 2D Neumann  Advection--Diffusion--Reaction''}\newline
$u_t=\beta_xu_{xx}+\beta_yu_{yy}-\alpha_xu_x-\alpha_yu_y+\gamma u+f(x,y)$\newline
where,\newline
$\beta_x=0.2$\newline
$\beta_y=0.1$\newline
$\alpha_x=2.0$\newline
$\alpha_y=3.0$\newline
$\gamma=5.0$\newline
$f=5(1-x-y)$
&
\textbf{BC:}\newline
\textit{left}: Neumann,\newline
$u_x(0,y,t)=1.0$\newline
\textit{right}: Neumann,\newline
$u_x(1,y,t)=1.0$\newline
\textit{bottom}: Neumann,\newline
$u_y(x,0,t)=1.0$\newline
\textit{top}: Neumann,\newline
$u_y(x,1,t)=1.0$\newline
\newline
\textbf{IC:}\newline
$u(x,y,0)=x+y$
&
$x\in[0,1]$\newline
$y\in[0,1]$\newline
$t\in[0,0.04]$\newline
\newline
$N_x=51$\newline
$N_y=51$\newline
$N_t=101$
&
$u(x,y,t)=x+y$
&
\mbox{} \newline
BE \newline
CN \newline
BDF2 \newline
SDIRK2 \newline
TR-BDF2 \newline
SSP-RK3 \newline
IMEX \newline
&
\textbf{t = 0.04:}\newline
$5.4 \times 10^{-3}$ \newline
$5.39 \times 10^{-3}$ \newline
$5.39 \times 10^{-3}$ \newline
$5.39 \times 10^{-3}$ \newline
$5.39 \times 10^{-3}$ \newline
$5.29 \times 10^{-3}$ \newline
$5.33 \times 10^{-3}$ \newline
&
\textbf{t = 0.04:}\newline
$3.72 \times 10^{-2}$ \newline
$3.72 \times 10^{-2}$ \newline
$3.72 \times 10^{-2}$ \newline
$3.72 \times 10^{-2}$ \newline
$3.72 \times 10^{-2}$ \newline
$3.67 \times 10^{-2}$ \newline
$3.69 \times 10^{-2}$ \newline \\

7 &
\textbf{``Steady 2D Mixed Tensor Diffusion''}\newline
$\beta_xu_{xx}+\beta_yu_{yy}+\eta_{xy}u_{yx}+\eta_{yx}u_{xy}+f(x,y)=0$\newline
where,\newline
$\beta_x=\beta_y=1.0$\newline
$\eta_{xy}=\eta_{yx}=0.5$\newline
$\alpha_x=\alpha_y=0.0$\newline
$\gamma=0.0$
&
\textbf{BC:}\newline
\textit{left}: Dirichlet,\newline
$u(0,y)=0$\newline
\textit{right}: Dirichlet,\newline
$u(1,y)=0$\newline
\textit{bottom}: Dirichlet,\newline
$u(x,0)=0$\newline
\textit{top}: Dirichlet,\newline
$u(x,1)=0$
&
$x\in[0,1]$\newline
$y\in[0,1]$\newline
\newline
$N_x=51$\newline
$N_y=51$
&
$u(x,y)=\sin(\pi x)\sin(\pi y)$
&
--
& 
$3.93 \times 10^{-4}$ \newline
&
$7.9 \times 10^{-4}$ \newline \\

\bottomrule
\end{tabularx}
\end{table}

\clearpage

\begin{table}[htbp]
\centering
\caption{Solver validation results for benchmark ODE/PDE cases, continued (2/3).}
\label{tab:validation_dataset_summary_continued}

\scriptsize
\setlength{\tabcolsep}{2.0pt}
\renewcommand{\arraystretch}{1.35}

\rowcolors{2}{gray!18}{white}
\begin{tabularx}{\linewidth}{@{}
>{\centering\arraybackslash}p{0.25cm}
>{\raggedright\arraybackslash}p{2.75cm}
>{\raggedright\arraybackslash}p{2.75cm}
>{\raggedright\arraybackslash}p{1.35cm}
L
>{\centering\arraybackslash}p{1.0cm}
>{\centering\arraybackslash}p{1.5cm}
>{\centering\arraybackslash}p{1.5cm}
@{}}
\toprule
\textbf{No.} &
\textbf{\makecell{Equation\\Type}} &
\textbf{\makecell{BC / IC \\Specification}} &
\textbf{\makecell{Domain}} &
\textbf{\makecell{Exact\\Solution}} &
\textbf{\makecell{Time\\Scheme}} &
\textbf{\makecell{$L_2$}} &
\textbf{$L_{\infty}$} \\
\midrule

8 &
\textbf{``Steady 2D Poisson Equation''}\newline
$\beta_xu_{xx}+\beta_yu_{yy}+f(x,y)=0$\newline
where,\newline
$\beta_x=\beta_y=1.0$\newline
$\alpha_x=\alpha_y=0.0$\newline
$\gamma=0.0$
&
\textbf{BC:}\newline
\textit{left}: Dirichlet,\newline
$u(0,y)=0$\newline
\textit{right}: Dirichlet,\newline
$u(1,y)=0$\newline
\textit{bottom}: Dirichlet,\newline
$u(x,0)=0$\newline
\textit{top}: Dirichlet,\newline
$u(x,1)=0$
&
$x\in[0,1]$\newline
$y\in[0,1]$\newline
\newline
$N_x=51$\newline
$N_y=51$
&
$u(x,y)=\sin(\pi x)\sin(\pi y)$
&
--
&
$4.87 \times 10^{-4}$ 
&
$9.86 \times 10^{-4}$ \\

9 &
\textbf{``Steady 2D Robin  Advection--Diffusion--Reaction''}\newline
$\beta_xu_{xx}+\beta_yu_{yy}+\eta_{xy}u_{yx}+\eta_{yx}u_{xy}-\alpha_xu_x-\alpha_yu_y+\gamma u+f(x,y)=0$\newline
where,\newline
$\beta_x=\beta_y=1.0$\newline
$\eta_{xy}=\eta_{yx}=0.5$\newline
$\alpha_x=\alpha_y=0.8$\newline
$\gamma=0.5$\newline
$f(x,y)=-1.9e^{x+y}$
&
\textbf{BC:}\newline
\textit{left}: Robin,\newline
$u(0,y)-1.5u(0,y)=-0.5e^y$\newline
\textit{right}: Robin,\newline
$u(1,y)+1.5u(1,y)=2.5e^{1+y}$\newline
\textit{bottom}: Robin,\newline
$u(x,0)-1.5u(x,0)=-0.5e^x$\newline
\textit{top}: Robin,\newline
$u(x,1)+1.5u(x,1)=2.5e^{x+1}$
&
$x\in[0,1]$\newline
$y\in[0,1]$\newline
\newline
$N_x=81$\newline
$N_y=81$
&
$u(x,y)=e^{x+y}$
&
--
&
$3.56 \times 10^{-4}$ 
&
$1.2 \times 10^{-3}$ \\

10 &
\textbf{``Transient 1D Heat Eigenmode''}\newline
$u_t=\beta u_{xx}$\newline
where,\newline
$\beta=0.04$
&
\textbf{BC:}\newline
\textit{left}: Dirichlet,\newline
$u(0,t)=0$\newline
\textit{right}: Dirichlet,\newline
$u(1,t)=0$\newline
\newline
\textbf{IC:}\newline
$u(x,0)=\sin(\pi x)$
&
$x\in[0,1]$\newline
$t\in[0,0.05]$\newline
\newline
$N_x=121$\newline
$N_t=41$
&
$u(x,t)=e^{-\beta\pi^2t}\sin(\pi x)$
&
CN
&
\textbf{t = 0:}\newline
$1.45 \times 10^{-16}$\newline
\textbf{t = 0.025:}\newline
$7.79 \times 10^{-5}$\newline
\textbf{t = 0.05:}\newline
$1.43 \times 10^{-4}$\newline
&
\textbf{t = 0:}\newline
$4.44 \times 10^{-16}$\newline
\textbf{t = 0.025:}\newline
$1.13 \times 10^{-4}$\newline
\textbf{t = 0.05:}\newline
$2.05 \times 10^{-4}$ \\

11 &
\textbf{``Transient 2D Mixed-BC Cross-Diffusion''}\newline
$u_t=\beta_xu_{xx}+\beta_yu_{yy}+\eta_{xy}u_{yx}+\eta_{yx}u_{xy}-\alpha_xu_x-\alpha_yu_y+\gamma u+f(x,y)$\newline
where,\newline
$\beta_x=0.02$,\newline 
$\beta_y=0.025$\newline
$\eta_{xy}=0.006$\newline
$\eta_{yx}=0.004$\newline
$\alpha_x=0.10$\newline
$\alpha_y=-0.08$\newline
$\gamma=0.05$
&
\textbf{BC:}\newline
\textit{left}: Dirichlet,\newline
$u(0,y,t)=1+y$\newline
\textit{bottom}: Dirichlet,\newline
$u(x,0,t)=1+x$\newline
\textit{top}: Neumann,\newline
$u_y(x,1,t)=1+x$\newline
\textit{right}: Robin,\newline
$1.5u(1,y,t)+0.014u_x(1,y,t)=3.014+3.014y$\newline
\newline
\textbf{IC:}\newline
$u(x,y,0)=1+x+y+xy$
&
$x\in[0,1]$\newline
$y\in[0,1]$\newline
$t\in[0,0.005]$\newline
\newline
$N_x=41$\newline
$N_y=41$\newline
$N_t=11$
&
$u(x,y,t)=1+x+y+xy$
&
SSP-RK3 \newline
\newline
\newline
\newline
\newline
\newline
\newline
IMEX \newline
\newline
\newline
\newline
&
\textbf{t = 0:}\newline
$3.12 \times 10^{-16}$\newline
\textbf{t = 0.0025:}\newline
$7.72 \times 10^{-6}$\newline
\textbf{t = 0.005:}\newline
$1.26 \times 10^{-5}$ \newline
\newline
\textbf{t = 0:}\newline
$3.12 \times 10^{-16}$\newline
\textbf{t = 0.0025:}\newline
$1.02 \times 10^{-5}$\newline
\textbf{t = 0.005:}\newline
$1.48 \times 10^{-5}$ 
&
\textbf{t = 0:}\newline
$8.88 \times 10^{-16}$\newline
\textbf{t = 0.025:}\newline
$4.4 \times 10^{-5}$\newline
\textbf{t = 0.05:}\newline
$7.7 \times 10^{-5}$ \newline 
\newline 
\textbf{t = 0:}\newline
$8.88 \times 10^{-16}$\newline
\textbf{t = 0.0025:}\newline
$6.37 \times 10^{-5}$\newline
\textbf{t = 0.005:}\newline
$1.03 \times 10^{-4}$ \\

12 &
\textbf{``Transient 2D Heat Eigenmode''}\newline
$u_t=\beta u_{xx}+\beta u_{yy}$\newline
where,\newline
$\beta=0.05$
&
\textbf{BC:}\newline
\textit{left}: Dirichlet,\newline
$u(0,y,t)=0$\newline
\textit{right}: Dirichlet,\newline
$u(1,y,t)=0$\newline
\textit{bottom}: Dirichlet,\newline
$u(x,0,t)=0$\newline
\textit{top}: Dirichlet,\newline
$u(x,1,t)=0$\newline
\newline
\textbf{IC:}\newline
$u(x,y,0)=\sin(\pi x)\sin(\pi y)$
&
$x\in[0,1]$\newline
$y\in[0,1]$\newline
$t\in[0,0.05]$\newline
\newline
$N_x=51$\newline
$N_y=51$\newline
$N_t=31$
&
$u(x,y,t)=e^{-2\beta\pi^2t}\sin(\pi x)$\newline
$\sin(\pi y)$
&
\mbox{} \newline
BE \newline
CN \newline
BDF2 \newline
SDIRK2 \newline
TR-BDF2 \newline
&
\textbf{t = 0.04:}\newline
$1.24 \times 10^{-5}$ \newline
$2.57 \times 10^{-5}$ \newline
$2.48 \times 10^{-5}$ \newline
$2.57 \times 10^{-5}$ \newline
$2.57 \times 10^{-5}$ \newline
&
\textbf{t = 0.04:}\newline
$2.62 \times 10^{-5}$ \newline
$4.64 \times 10^{-5}$ \newline
$4.45 \times 10^{-5}$ \newline
$4.64 \times 10^{-5}$ \newline
$4.64 \times 10^{-5}$ \newline \\

14 &
\textbf{``Transient 2D Neumann Advection--Diffusion''}\newline
$u_t=\beta_xu_{xx}+\beta_yu_{yy}-\alpha_xu_x-\alpha_yu_y+\gamma u+f$\newline
where,\newline
$\beta_x=1.0$\newline
$\beta_y=0.5$\newline
$\alpha_x=2.0$\newline
$\alpha_y=3.0$\newline
$\gamma=0.0$\newline
$f=6$
&
\textbf{BC:}\newline
\textit{left}: Neumann,\newline
$u_x(0,y,t)=1.0$\newline
\textit{right}: Neumann,\newline
$u_x(1,y,t)=1.0$\newline
\textit{bottom}: Neumann,\newline
$u_y(x,0,t)=1.0$\newline
\textit{top}: Neumann,\newline
$u_y(x,1,t)=1.0$\newline
\newline
\textbf{IC:}\newline
$u(x,y,0)=x+y$
&
$x\in[0,1]$\newline
$y\in[0,1]$\newline
$t\in[0,1]$\newline
\newline
$N_x=51$\newline
$N_y=51$\newline
$N_t=5$
&
$u(x,y,t)=x+y+t$
&
BDF2
&
$2.94 \times 10^{-3}$
&
$6.18 \times 10^{-3}$ \\

\bottomrule
\end{tabularx}
\end{table}

\clearpage

\begin{table}[htbp]
\centering
\caption{Solver validation results for benchmark ODE/PDE cases, continued (3/3).}
\label{tab:validation_dataset_summary_continued_2}

\scriptsize
\setlength{\tabcolsep}{2.0pt}
\renewcommand{\arraystretch}{1.35}

\rowcolors{2}{gray!18}{white}
\begin{tabularx}{\linewidth}{@{}
>{\centering\arraybackslash}p{0.25cm}
>{\raggedright\arraybackslash}p{2.75cm}
>{\raggedright\arraybackslash}p{2.75cm}
>{\raggedright\arraybackslash}p{1.35cm}
L
>{\centering\arraybackslash}p{1.0cm}
>{\centering\arraybackslash}p{1.5cm}
>{\centering\arraybackslash}p{1.5cm}
@{}}
\toprule
\textbf{No.} &
\textbf{\makecell{Equation\\Type}} &
\textbf{\makecell{BC / IC \\Specification}} &
\textbf{\makecell{Domain}} &
\textbf{\makecell{Exact\\Solution}} &
\textbf{\makecell{Time\\Scheme}} &
\textbf{\makecell{$L_2$}} &
\textbf{$L_{\infty}$} \\
\midrule

14 &
\textbf{``Transient 2D Decaying  Advection--Diffusion--Reaction''}\newline
$u_t=\beta_xu_{xx}+\beta_yu_{yy}-\alpha_xu_x-\alpha_yu_y+\gamma u+f(x,y,t)$\newline
where,\newline
$\beta_x=\beta_y=0.05$\newline
$\alpha_x=0.3$\newline
$\alpha_y=0.2$\newline
$\gamma=-2.0$
&
\textbf{BC:}\newline
\textit{left}: Dirichlet,\newline
$u(0,y,t)=0$\newline
\textit{right}: Dirichlet,\newline
$u(1,y,t)=0$\newline
\textit{bottom}: Dirichlet,\newline
$u(x,0,t)=0$\newline
\textit{top}: Dirichlet,\newline
$u(x,1,t)=0$\newline
\newline
\textbf{IC:}\newline
$u(x,y,0)=\sin(\pi x)\sin(\pi y)$
&
$x\in[0,1]$\newline
$y\in[0,1]$\newline
$t\in[0,1]$\newline
\newline
$N_x=51$\newline
$N_y=51$\newline
$N_t=601$
&
$u(x,y,t)=e^{-2t}\sin(\pi x)$\newline
$\sin(\pi y)$
&
BE \newline
\newline
\newline
\newline
\newline
\newline
\newline
CN \newline
\newline
\newline
\newline
\newline
\newline
\newline
BDF2 \newline
\newline
\newline
\newline
\newline
\newline
\newline
SDIRK2 \newline
\newline
\newline
\newline
\newline
\newline
\newline
TR-BDF2 \newline
\newline
\newline
\newline
\newline
\newline
\newline
SSP-RK3 \newline
\newline
\newline
\newline
\newline
\newline
\newline
IMEX \newline
\newline
\newline
\newline
\newline
\newline
\newline
&
\textbf{t = 0:}\newline
$2.09 \times 10^{-16}$\newline
\textbf{t = 0.5:}\newline
$4.50 \times 10^{-4}$\newline
\textbf{t = 1:}\newline
$1.99 \times 10^{-4}$ \newline
\newline
\textbf{t = 0:}\newline
$2.09 \times 10^{-16}$\newline
\textbf{t = 0.5:}\newline
$6.18 \times 10^{-4}$\newline
\textbf{t = 1:}\newline
$3.13 \times 10^{-4}$ \newline
\newline
\textbf{t = 0:}\newline
$2.09 \times 10^{-16}$\newline
\textbf{t = 0.5:}\newline
$5.74 \times 10^{-4}$\newline
\textbf{t = 1:}\newline
$2.73 \times 10^{-4}$ \newline
\newline
\textbf{t = 0:}\newline
$2.09 \times 10^{-16}$\newline
\textbf{t = 0.5:}\newline
$6.18 \times 10^{-4}$\newline
\textbf{t = 1:}\newline
$3.12 \times 10^{-4}$ \newline
\newline
\textbf{t = 0:}\newline
$2.09 \times 10^{-16}$\newline
\textbf{t = 0.5:}\newline
$5.89 \times 10^{-4}$\newline
\textbf{t = 1:}\newline
$2.87 \times 10^{-4}$ \newline
\newline
\textbf{t = 0:}\newline
$2.09 \times 10^{-16}$\newline
\textbf{t = 0.5:}\newline
$4.58 \times 10^{-4}$\newline
\textbf{t = 1:}\newline
$2.14 \times 10^{-4}$ \newline
\newline
\textbf{t = 0:}\newline
$2.09 \times 10^{-16}$\newline
\textbf{t = 0.5:}\newline
$6.95 \times 10^{-4}$\newline
\textbf{t = 1:}\newline
$3.68 \times 10^{-4}$ \newline
&
\textbf{t = 0:}\newline
$5.55 \times 10^{-16}$\newline
\textbf{t = 0.5:}\newline
$8.83 \times 10^{-4}$\newline
\textbf{t = 1:}\newline
$4.03 \times 10^{-4}$ \newline
\newline 
\textbf{t = 0:}\newline
$5.55 \times 10^{-16}$\newline
\textbf{t = 0.5:}\newline
$1.34 \times 10^{-3}$\newline
\textbf{t = 1:}\newline
$7.09 \times 10^{-4}$ \newline
\newline 
\textbf{t = 0:}\newline
$5.55 \times 10^{-16}$\newline
\textbf{t = 0.5:}\newline
$1.25 \times 10^{-3}$\newline
\textbf{t = 1:}\newline
$6.34 \times 10^{-4}$ \newline
\newline
\textbf{t = 0:}\newline
$5.55 \times 10^{-16}$\newline
\textbf{t = 0.5:}\newline
$1.33 \times 10^{-3}$\newline
\textbf{t = 1:}\newline
$7.09 \times 10^{-4}$ \newline
\newline
\textbf{t = 0:}\newline
$5.55 \times 10^{-16}$\newline
\textbf{t = 0.5:}\newline
$1.28 \times 10^{-3}$\newline
\textbf{t = 1:}\newline
$6.63 \times 10^{-4}$ \newline
\newline
\textbf{t = 0:}\newline
$5.55 \times 10^{-16}$\newline
\textbf{t = 0.5:}\newline
$1.01 \times 10^{-3}$\newline
\textbf{t = 1:}\newline
$5.02 \times 10^{-4}$ \newline
\newline
\textbf{t = 0:}\newline
$5.55 \times 10^{-16}$\newline
\textbf{t = 0.5:}\newline
$1.46 \times 10^{-3}$\newline
\textbf{t = 1:}\newline
$8.04 \times 10^{-4}$ \newline
\\

\bottomrule
\end{tabularx}
\end{table}


\subsection{Input-Handling Graph Details}
\label{app:input_handling}

\subsubsection{Graph State}
\label{app:graph_state}

The input-handling graph keeps a structured state object across turns. Table~\ref{tab:graph_state_fields} lists the main fields in this state and their purpose.

\begin{table}[htbp]
\centering
\caption{State fields used by the input-handling graph.}
\label{tab:graph_state_fields}
\small
\renewcommand{\arraystretch}{1.16}
\begin{tabular}{p{0.28\linewidth} p{0.62\linewidth}}
\toprule
\textbf{State field} & \textbf{Purpose} \\
\midrule

Latest user message &
Stores the current user message \\

Message classification &
Stores the Orchestrator output, including the intent label and the split between conversational content and specification-bearing content. \\

Response status &
Records whether the latest agent call returned a usable response, failed or produced an invalid format. This field is used for error handling. \\

Conversation history &
Stores the full user-agent interaction as fallback context. The full history is not normally passed to agents directly. \\

Recent turns &
Stores the last few turns passed to agents as local context. In the current implementation, the default window size is four messages. \\

Memory summary &
Stores a compact semantic summary of the conversation between the user and the interviewer agent. It preserves older conversation history in a concise format. It is produced by the Summarizer and passed to later agents. \\

Complete specification &
Stores the current canonical ODE/PDE specification. This is the object validated for execution and finalised for downstream use. \\

Candidate patch &
Stores the incremental JSON patch produced by the Extractor from the current specification-bearing text. The Extractor edits only the changed fields rather than rebuilding the full specification. \\

Correction log &
Stores meaningful user changes and overrides, as identified by the Comparator. A change is treated as a correction only when it modifies earlier state. \\

Validation report &
Stores the Validator output, including patch validity, missing fields, warnings and suggestions used by the Repairer or Interviewer. \\

Repair retry count &
Stores how many repair attempts have been made for the current invalid patch. The counter is reset after a valid patch is accepted. \\

Maximum repair retries &
Stores the upper bound on repair attempts before the system asks the user for clarification. We set this value to two in the current implementation. \\

\bottomrule
\end{tabular}
\end{table}

The state is designed around one invariant: the complete specification changes only after a candidate patch has passed validation. Conversation, clarification and repair can update auxiliary fields, but they do not silently alter the executable problem definition. This invariant is the main reason we separate candidate patches from the complete specification.

\subsubsection{Agent and Tool Responsibilities}
\label{app:agent_tool_roles}

Table~\ref{tab:agent_prompt_overview} summarises the responsibilities of the agents and tools in the input-handling graph. Together, these components classify user intent, extract incremental patches, validate updates, repair invalid patches when possible, and finalise executable ODE/PDE specifications.

\begin{table}[htbp]
\centering
\caption{Roles and responsibilities of the input-handling agents and tools.}
\label{tab:agent_prompt_overview}
\small
\renewcommand{\arraystretch}{1.18}
\setlength{\tabcolsep}{6pt}
\begin{tabularx}{0.95\linewidth}{
    >{\centering\arraybackslash}p{0.08\linewidth}
    >{\raggedright\arraybackslash}p{0.24\linewidth}
    >{\raggedright\arraybackslash}X
}
\toprule
\rowcolor{gray!12}
\textbf{No.} & \textbf{Agent / Tool} & \textbf{Responsibility} \\
\midrule

1 &
Summarizer Tool &
Condenses older dialogue into a compact memory summary. The summary keeps user-provided modelling facts, corrections and unresolved requirements, while removing repeated conversational detail. \\

2 &
Orchestrator Agent &
Classifies the latest user message as conversation, specification, mixed input, assistant-generated specification or no usable input. It also separates conversational text from executable problem content before routing. \\

3 &
Conversation Agent &
Answers ordinary dialogue without directly editing the complete specification. When the message contains both dialogue and executable content, it forwards only the specification-bearing part to extraction. For assistant-generated specification, it rewrites vague scientific requests into concrete mathematical content before extraction. \\

4 &
Interviewer Agent &
Asks clarification questions when the current specification is incomplete, ambiguous or failed validation. The agent asks user-facing questions about missing equations, coefficients, domains, boundary conditions, initial conditions, solver settings or training settings. \\

5 &
Extractor Agent &
Converts specification-bearing text into JSON patch operations. The prompt is problem-type specific for PDEs and ODEs, includes schema examples and asks for incremental add, replace or remove operations rather than a full schema rewrite. \\

6 &
Comparator Tool &
Compares the proposed patch with the current complete specification and records added, modified and removed fields. This change log makes user corrections visible across turns. \\

7 &
Validator Agent &
Checks whether the proposed patch is syntactically valid, schema-consistent, path-correct and sufficient for the selected ODE/PDE class. It returns a structured validation report with status, missing conditions, warnings and repair suggestions, but does not modify the patch. \\

8 &
Repairer Agent &
Repairs invalid patches using the current complete specification, proposed patch and Validator feedback. The repair prompt focuses on formatting, normalisation and path errors. A repaired patch must pass validation before it can be applied. \\

9 &
Patcher Tool &
Applies a validated JSON patch to the complete specification. Failed patches and unrepaired patches do not alter the executable problem state. \\

10 &
Finalizer Agent &
Marks a completed ODE/PDE specification as ready, sets schema metadata, normalises the problem name and prepares the final JSON for dataset generation, solver execution and operator training. \\

\bottomrule
\end{tabularx}
\end{table}

\subsubsection{Validator Completion Criteria}
\label{app:completion_criteria}

The Validator agent marks a specification as complete only when the minimum executable content for the selected problem class is present. These checks are tied to the supported solver and training pipeline, rather than to a general mathematical definition of an ODE or PDE. Table~\ref{tab:input_completion_criteria} lists the minimum executable elements required by the Validator for each supported problem class prior to finalisation.

\begin{table}[htbp]
\centering
\caption{Completion criteria used by the Validator before finalisation.}
\label{tab:input_completion_criteria}
\small
\begin{tabular}{p{0.24\linewidth} p{0.66\linewidth}}
\toprule
\textbf{Problem class} & \textbf{Minimum executable specification} \\
\midrule

Transient 2D &
A time-dependent PDE with the dependent variable, two spatial coordinates, time variable, rectangular grid, coefficients or source terms, initial condition over the grid, boundary conditions on all four sides, time interval and solver or training settings. \\

Transient 1D &
A time-dependent PDE with the dependent variable, one spatial coordinate, time variable, one-dimensional grid, coefficients or source terms, initial condition over the grid, boundary conditions at both endpoints, time interval and solver or training settings. \\

Steady-state 2D &
A static PDE with the dependent variable, two spatial coordinates, rectangular grid, coefficients or source terms, boundary conditions on all four sides and solver or training settings. No initial condition or time interval is required. \\

Steady-state 1D &
A static differential equation with the dependent variable, one spatial coordinate, one-dimensional grid, coefficients or source terms, boundary conditions at the required endpoints and solver or training settings. No initial condition or time interval is required. \\

ODE & 
An ODE with state variables, an independent variable, parameters or forcing terms, an initial value, an integration interval or evaluation points, and solver or training settings. \\

\bottomrule
\end{tabular}
\end{table}

\subsubsection{Validation Scenarios}
\label{app:validation_scenarios}

The validation runners use scripted multi-turn scenarios. Each scenario contains a problem family, a difficulty label and an ordered user conversation with intent annotations. We report the examples as scenario cards rather than raw JSON. The runner uses the same turn order; the card format is included only to make the tested interaction readable.

The selected cards cover cases that occur during natural-language problem entry: delayed specification, unsupported mathematical forms, assistant-generated conversion, boundary-condition alignment and conversational interference. These cases stress different parts of the graph. Missing coefficients test the Validator and Interviewer agents. Unsupported wave or algebraic requests test the Conversation Agent's conversion path. Multi-turn boundary edits test whether the patch mechanism preserves earlier choices while applying a local change.

\newpage
\subsubsection*{\underline{\textbf{SCENARIO 1: Hard 1D Transient PDE with a Missing Coefficient}}}

\noindent\textbf{Problem family:} 1D PDE. 
\textbf{Difficulty:} H.

\begin{enumerate}[leftmargin=*, label=\textbf{Turn \arabic*:}]
    \item \textbf{Intent: specification.}
    Define a transient 1D PDE on \texttt{x in [0, L]} with \texttt{L = 1.35} and \texttt{t in [0, T]} with \texttt{T = 1.05}. Use 74 spatial grid points and 70 temporal grid points.

    \item \textbf{Intent: specification.}
    Use the equation string \texttt{u\_t = a*u\_xx + d*u\_x + f*u + g}, but leave \texttt{a} unspecified. Set \texttt{d(x)=n*x}, where \texttt{n} is sampled uniformly from \texttt{[0.1, 0.2]}. Set \texttt{f(x)=alpha*x}, where \texttt{alpha} is sampled uniformly from \texttt{[0.01, 0.03]}.

    \item \textbf{Intent: conversation.}
    Ask what happens when a coefficient is missing.

    \item \textbf{Intent: specification.}
    Supply the missing coefficient with \texttt{a = 0.1}. Set the source string to \texttt{g(x)=beta*x\string^2 + gamma*sin(pi*x/L)}, where \texttt{beta} is sampled uniformly from \texttt{[0.01, 0.03]} and \texttt{gamma} is sampled uniformly from \texttt{[-0.02, 0.02]}.

    \item \textbf{Intent: specification.}
    Set the initial condition string to \texttt{sin(pi*x/L) + 0.1*sin(4*pi*x/L)}. Use boundary conditions \texttt{u(0,t)=0} and \texttt{u(L,t)=BL}, where \texttt{BL} is sampled uniformly from \texttt{[-0.05, 0.05]}. Use 2 training samples and 2 testing samples.
\end{enumerate}

\subsubsection*{\underline{\textbf{SCENARIO 2: Advanced-Hard ODE with an Unsupported Initial Request}}}

\noindent\textbf{Problem family:} ODE. 
\textbf{Difficulty:} AH.

\begin{enumerate}[leftmargin=*, label=\textbf{Turn \arabic*:}]
    \item \textbf{Intent: specification.}
    Request an unsupported algebraic relation \texttt{u = 2 + t} with a sinusoidal source.

    \item \textbf{Intent: conversation.}
    Ask for the difference between an algebraic equation and an ODE specification.

    \item \textbf{Intent: assistant generated specification.}
    Ask the system to generate and apply a supported ODE version using a forcing term.

    \item \textbf{Intent: specification.}
    Set the supported equation string to \texttt{u\_t = f*u + g(t)}. Use \texttt{t in [0, 1.2]} and 90 temporal grid points. Set \texttt{f = -0.05}.

    \item \textbf{Intent: specification.}
    Set the forcing string to \texttt{g(t) = 0.2 + 0.8 + G*sin(pi*t/T)}, where \texttt{G} is sampled uniformly from \texttt{[-0.1, 0.1]}. Remove the invalid time-varying initial condition, set \texttt{u(0) = 2}, use no boundary conditions and use 4 training samples and 4 testing samples.
\end{enumerate}

\subsubsection*{\underline{\textbf{SCENARIO 3: Easy 2D Transient PDE}}}

\noindent\textbf{Problem family:} 2D PDE. 
\textbf{Difficulty:} E.

\begin{enumerate}[leftmargin=*, label=\textbf{Turn \arabic*:}]
    \item \textbf{Intent: specification.}
    Set up a transient 2D PDE on \texttt{x in [-0.2, 1.2]}, \texttt{y in [0, 1.4]} and \texttt{t in [0, 1.1]}. Use 64 grid points in \texttt{x}, 64 grid points in \texttt{y} and 64 time points. Use the equation string \texttt{u\_t = a*u\_xx + b*u\_yy + f*u + g}. Sample \texttt{a} and \texttt{b} uniformly from \texttt{[0.05, 0.2]}, sample \texttt{f} uniformly from \texttt{[-0.08, -0.01]} and set the source string to \texttt{g(x,y)=gamma*(x + 0.3 - 0.1)*y}, where \texttt{gamma} is sampled uniformly from \texttt{[-0.05, 0.05]}.

    \item \textbf{Intent: specification.}
    Set the initial condition string to \texttt{sin(pi*(x + 0.3 - 0.1)/1.4)*sin(pi*y/L\_y) + 0.15*sin(2*pi*(x + 0.3 - 0.1)/1.4)*sin(3*pi*y/L\_y)}. Use Dirichlet boundary conditions: \texttt{u=0 on x=-0.2}, \texttt{u=beta\_x*sin(pi*y/L\_y) on x=L\_x}, \texttt{u=0 on y=0} and \texttt{u=beta\_y*sin(pi*(x + 0.3 - 0.1)/1.4) on y=L\_y}. Sample \texttt{beta\_x} and \texttt{beta\_y} uniformly from \texttt{[-0.05, 0.05]}. Use 2 training samples and 2 testing samples.
\end{enumerate}

\subsubsection*{\underline{\textbf{SCENARIO 4: Advanced-Hard 2D PDE with an Unsupported Wave Equation Request}}}

\noindent\textbf{Problem family:} 2D PDE. 
\textbf{Difficulty:} AH.

\begin{enumerate}[leftmargin=*, label=\textbf{Turn \arabic*:}]
    \item \textbf{Intent: specification.}
    Request a 2D wave equation with \texttt{u\_tt} on \texttt{x in [0, 1.2]}, \texttt{y in [0, 1.4]} and \texttt{t in [0, 1]}.

    \item \textbf{Intent: conversation.}
    Ask whether second-order time equations such as \texttt{u\_tt} are supported.

    \item \textbf{Intent: assistant generated specification.}
    Ask the system to apply the closest supported transient heat equation instead.

    \item \textbf{Intent: specification.}
    Set the supported equation string to \texttt{u\_t = a*u\_xx + b*u\_yy + f*u}, with \texttt{a = 0.1}, \texttt{b = 0.1} and \texttt{f = -0.02}. Use \texttt{x in [0, 1.2]}, \texttt{y in [0, 1.4]} and \texttt{t in [0, 1]}. Use 70 grid points in \texttt{x}, 90 grid points in \texttt{y} and 90 temporal grid points.

    \item \textbf{Intent: specification.}
    Set the initial condition string to \texttt{sin(pi*x/L\_x)*sin(pi*y/L\_y) + 0.1*sin(2*pi*x/L\_x)*sin(3*pi*y/L\_y)}. Use homogeneous Dirichlet boundary conditions on all four edges, \texttt{u=0}. Use 4 training samples and 4 testing samples.
\end{enumerate}

\subsubsection*{\underline{\textbf{SCENARIO 5: Medium 1D Advection PDE}}}

\noindent\textbf{Problem family:} 1D PDE. 
\textbf{Difficulty:} M.

\begin{enumerate}[leftmargin=*, label=\textbf{Turn \arabic*:}]
    \item \textbf{Intent: specification.}
    Create a transient 1D advection PDE on \texttt{x in [-0.3, L]} with \texttt{L = 1.3} and \texttt{t in [0, T]} with \texttt{T = 0.85}. Use 78 spatial grid points and 82 temporal grid points.

    \item \textbf{Intent: specification.}
    Set the equation string to \texttt{u\_t = d*u\_x + g}. Use the coefficient string \texttt{d(x)=-(0.3 + 0.1*cos(2*pi*(x + 0.3)/1.6))}.

    \item \textbf{Intent: specification.}
    Use the source string \texttt{g(x)=gamma*sin(pi*(x + 0.3)/1.6)}, where \texttt{gamma} is sampled uniformly from \texttt{[-0.05, 0.05]}. Set the initial condition string to \texttt{sin(pi*(x + 0.3)/1.6) + 0.15*cos(3*pi*(x + 0.3)/1.6)}. Use inflow at \texttt{x=-0.3} with \texttt{u(-0.3,t)=B0}, where \texttt{B0} is sampled uniformly from \texttt{[-0.05, 0.05]}, open outflow at \texttt{x=L} and 2 training samples and 2 testing samples.
\end{enumerate}

\subsection{Example Chat Histories and Generated JSON Specifications}
\label{sec:chat_history_json_examples}

\begin{figure*}[h]
    \centering
    \includegraphics[width=\textwidth]{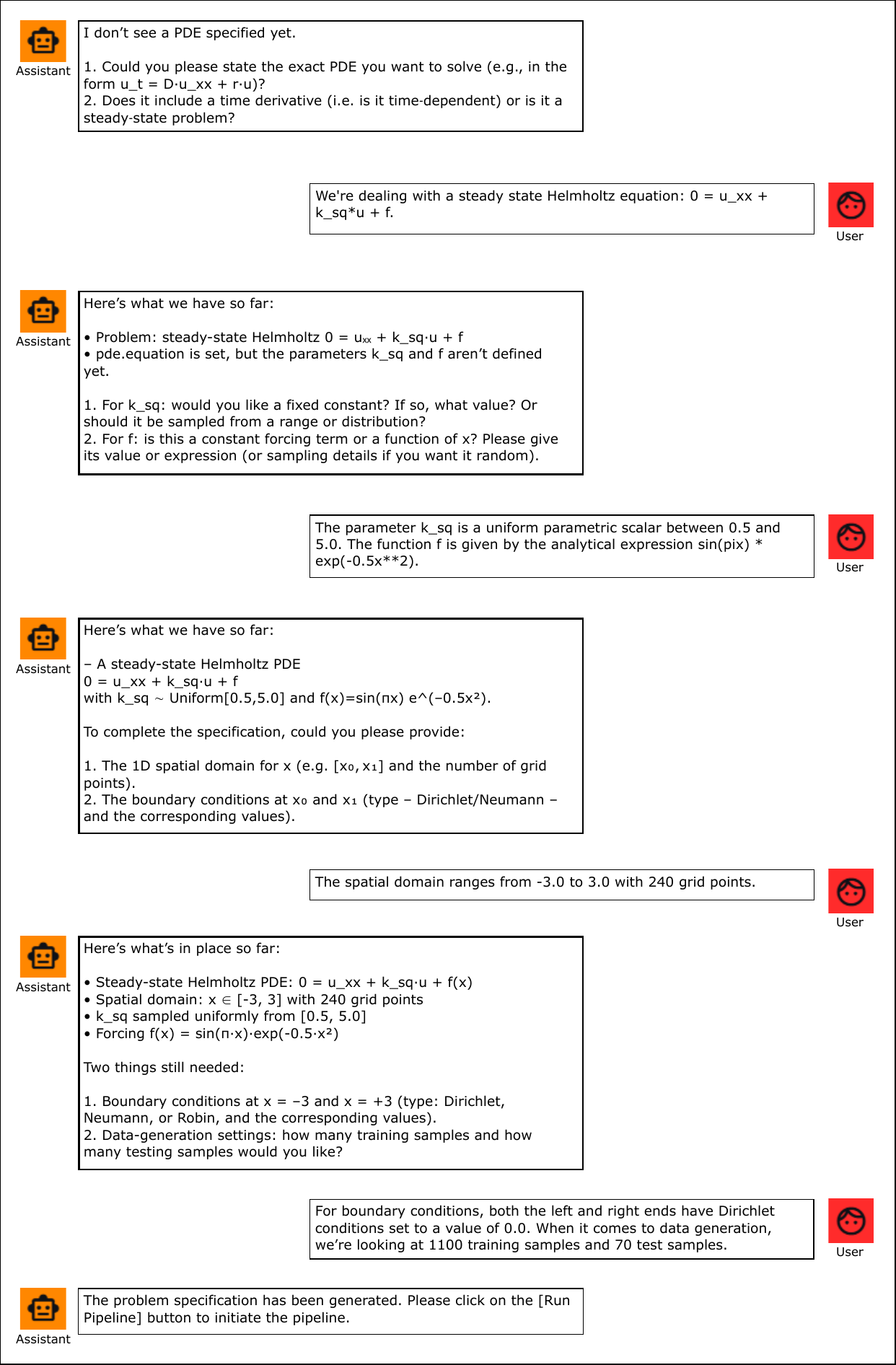}
    \caption{Chat history for generating a steady 1D parametric Helmholtz specification.}
    \label{fig:chat_history_1}
\end{figure*}

\begin{figure*}[h]
    \centering
    \includegraphics[width=\textwidth]{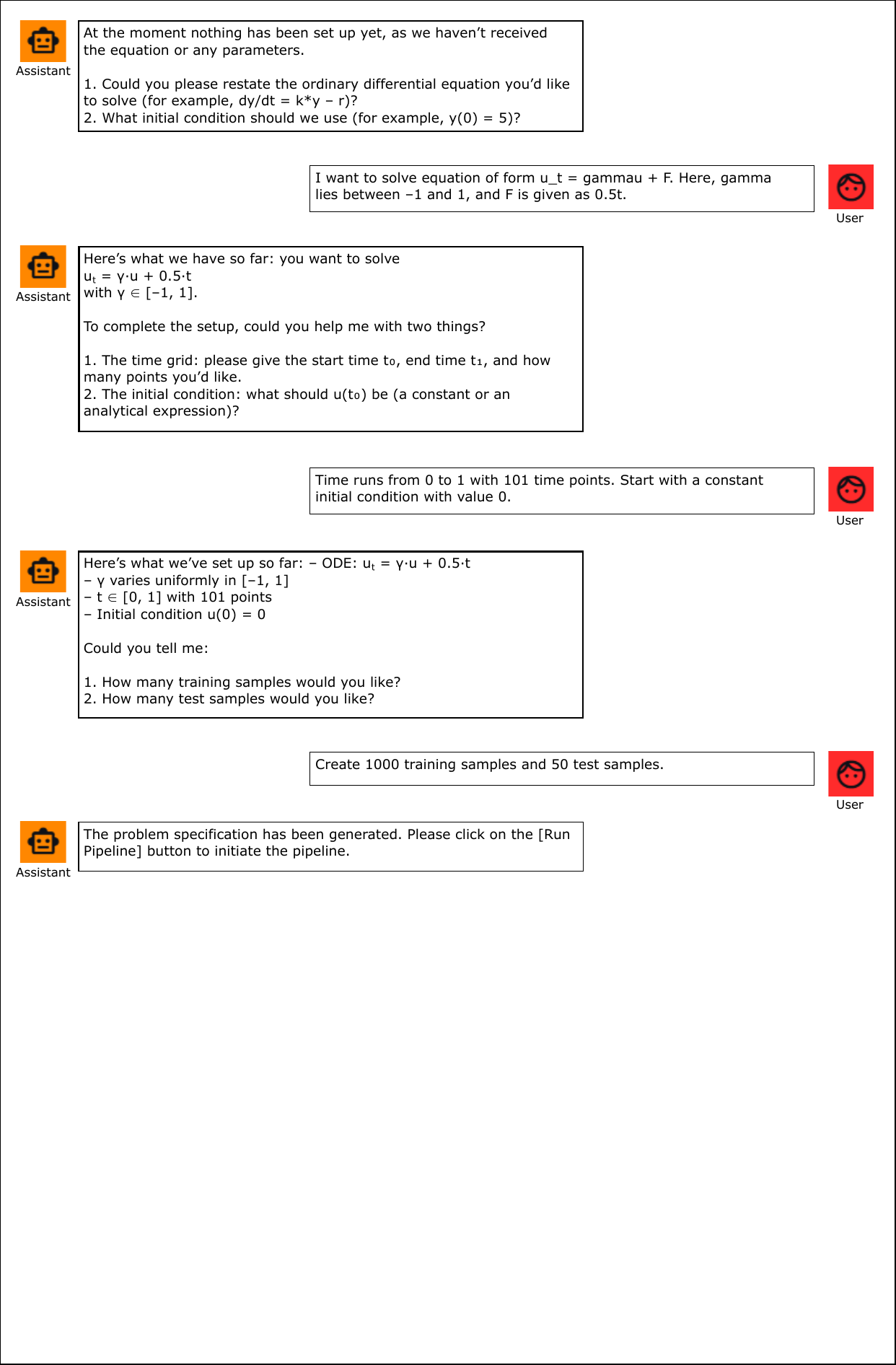}
    \caption{Chat history for generating a parametric exponential-growth ODE specification.}
    \label{fig:chat_history_2}
\end{figure*}

This section gives two representative examples of how user interactions are converted into executable problem specifications. Figures~\ref{fig:chat_history_1} and~\ref{fig:chat_history_2} show the chat histories for a steady 1D Helmholtz problem and a temporal ODE problem, respectively. The corresponding JSON blocks show the final structured specifications passed to the data-generation and solver modules. These examples illustrate how the same specification format can encode both spatial PDE cases and pure temporal ODE cases.

\definecolor{codebg}{RGB}{248,248,248}

\newtcblisting{jsonpagebox}[1][]{
  enhanced,
  width=\textwidth,
  height=\textheight,
  colback=codebg,
  colframe=black!60,
  arc=4mm,
  boxrule=0.5pt,
  title={#1},
  fonttitle=\bfseries,
  listing only,
  listing engine=listings,
  top=2mm,
  bottom=2mm,
  left=2mm,
  right=2mm,
  listing options={
    basicstyle=\ttfamily\fontsize{8.8}{13.8}\selectfont,
    breaklines=true,
    columns=fullflexible,
    keepspaces=true,
    showstringspaces=false
  }
}

\clearpage
\noindent
\begin{jsonpagebox}[Generated JSON Specification: Steady 1D Parametric Helmholtz Problem]
{
  "domain": {
    "x": [-3.0, 3.0, 240]
  },

  "pde": {
    "equation": "0 = u_xx + k_sq*u + f",
    "equation_type": "steady_1d_helmholtz",
    "parameters": {
      "k_sq": {
        "type": "parametric_scalar",
        "distribution": "uniform",
        "range": [0.5, 5.0]
      },

      "f": {
        "type": "analytical",
        "expression": "sin(pi*x) * exp(-0.5*x**2)"
      }
    }
  },

  "boundary_conditions": {
    "left": {
      "type": "dirichlet",
      "value": 0.0
    },
    "right": {
      "type": "dirichlet",
      "value": 0.0
    }
  },

  "data_generation": {
    "num_train_samples": 1100,
    "num_test_samples": 70
  }
}
\end{jsonpagebox}

\clearpage
\noindent
\begin{jsonpagebox}[Generated JSON Specification: Parametric Scalar Exponential Growth ODE]
{
  "domain": {},

  "time": [0.0, 1.0, 101],

  "pde": {
    "equation": "u_t = gamma*u + F",
    "equation_type": "parametric_scalar_growth_linear_source",
    "parameters": {
      "gamma": {
        "type": "parametric_scalar",
        "range": [-1, 1]
      },
      "F": { 
        "type": "analytical", 
        "expression": 0.5*t 
      }
    }
  },

  "initial_condition": {
    "type": "constant",
    "value": 0.0
  },

  "data_generation": {
    "num_train_samples": 1000,
    "num_test_samples": 50
  }
}
\end{jsonpagebox}

\clearpage

\end{document}